\documentclass[preprint,12pt]{elsarticle}

\usepackage{amssymb}
\usepackage{amsmath}

\usepackage{moreverb,url}
\usepackage[colorlinks,bookmarksopen,bookmarksnumbered,citecolor=red,urlcolor=red]{hyperref}
\usepackage{amsmath,amssymb,amsfonts}
\usepackage{graphicx}
\usepackage{textcomp}
\usepackage{subcaption}
\usepackage{multicol}
\usepackage{multirow}
\usepackage{xcolor}
\usepackage[ruled,lined,linesnumbered]{algorithm2e}
\usepackage{algorithmic}
\usepackage{algorithm2e} 
\usepackage{array}
\usepackage{stfloats}
\usepackage{url}
\usepackage{verbatim}
\usepackage{balance}
\usepackage{booktabs}

\newcommand{\ar}[1]{{\color{black}#1}} 
\newcommand{\cd}[1]{{\color{black}#1}} 
\newcommand{\rsf}[1]{{\color{black}#1}} 
\newcommand{\rev}[1]{{\color{black}#1}}

\journal{Computer Communications}

\begin{document}

\begin{frontmatter}



\title{Enhancing Cellular-enabled Collaborative Robots Planning through GNSS data for SAR Scenarios\tnoteref{a1}}

\tnotetext[a1]{Funding: The research leading to these results has been supported by the Spanish Ministry of Economic Affairs and Digital Transformation and the European Union – NextGeneration EU, in the framework of the Recovery Plan, Transformation and Resilience (PRTR) (Call UNICO I+D 5G 2021, ref. number TSI-063000-2021-6 and TSI-063000-2021-14), by the SNS JU under the European Union’s Horizon Europe research and innovation programme under Grant Agreement No. 101192521 (MultiX), by the CERCA Programme from the Generalitat de Catalunya and by the Project PID2024-157729OB-I00 funded by MICIU/AEI/10.13039/501100011033.}


\author[i2cat,upc]{Arnau Romero \corref{cor1}} 
\ead{arnau.romero@i2cat.net}
\author[i2cat]{Carmen Delgado} 
\author[upf]{Jana Baguer} 
\author[upc]{Ra\'ul Su\'arez} 
\author[i2cat,nec,icrea]{Xavier~Costa-P\'erez}

\cortext[cor1]{Corresponding author.}
\affiliation[i2cat]{organization={AI-driven Systems, i2CAT Foundation},
            city={Barcelona},
            country={Spain}}
\affiliation[upc]{organization={Universitat Polit\`{e}cnica de Catalunya, UPC},
            city={Barcelona},
            country={Spain}}
\affiliation[upf]{organization={Pompeu Fabra University, UPF},
            city={Barcelona},
            country={Spain}}
\affiliation[nec]{organization={NEC Laboratories Europe GmbH},
            city={Heidelberg},
            country={Germany}}
\affiliation[icrea]{organization={Institut Català de Recerca i Estudis Avançats, ICREA},
            city={Barcelona},
            country={Spain}}

\begin{abstract}
Cellular-enabled collaborative robots are becoming paramount in Search-and-Rescue (SAR) and emergency response. Crucially dependent on resilient mobile network connectivity, they serve as invaluable assets for tasks like rapid victim localization and the exploration of hazardous, otherwise unreachable areas. However, their reliance on battery power and the need for persistent, low-latency communication limit operational time and mobility. To address this, and considering the evolving capabilities of 5G/6G networks, we propose a novel SAR framework that includes Mission Planning and Mission Execution phases and that optimizes robot deployment. By considering parameters such as the exploration area size, terrain elevation, robot fleet size, communication-influenced energy profiles, desired exploration rate, and target response time, our framework determines the minimum number of robots required and their optimal paths to ensure effective coverage and timely data backhaul over mobile networks. Our results demonstrate the trade-offs between number of robots, explored area, and response time for wheeled and quadruped robots. Further, we quantify the impact of terrain elevation data on mission time and energy consumption, showing the benefits of incorporating real-world environmental factors that might also affect mobile signal propagation and connectivity into SAR planning. This framework provides critical insights for leveraging next-generation mobile networks to enhance autonomous SAR operations.
\end{abstract}

\begin{graphicalabstract}
\label{fig:graphical_abstract}
\hspace{-1.6cm}
\includegraphics[width=1.2\textwidth]{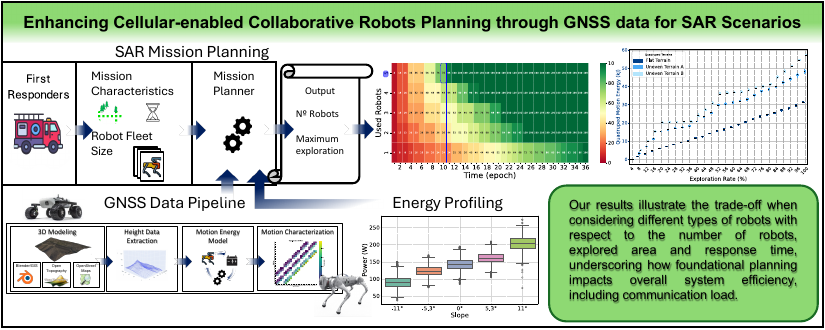}
\end{graphicalabstract}


\end{frontmatter}



\section{Introduction}

In mission-critical Search-and-Rescue (SAR) operations, rapid response to disasters and emergencies is paramount for saving lives. While human first responders bravely face these situations, their risks can be significantly mitigated by deploying mobile robotic fleets for victim localization and hazardous area exploration~\cite{SAR_UGR_2022}.
For these robots to be truly effective, robust and pervasive mobile network connectivity isn't just an enabler, it's a fundamental requirement. 
This connectivity facilitates real-time data exchange and effective coordination of multi-robot operations, which is essential to meet the rapid response requirements of SAR operations, optimizing zone exploration and task allocation while avoiding redundancy~\cite{Cao2022}. This coordination is often managed by a centralized task planner on an edge server~\cite{LNORM_2023} that collects robot feedbacks, maps explored areas, and generates optimal path plans for each robot.

However, mobile robots face a significant challenge in the form of energy constraints since they normally rely solely on batteries. Increasing battery capacity would lead to added weight, resulting in higher mobility energy consumption and thus, a design trade-off~\cite{Albonico2021}. Extensive research has \cd{focused on} energy-efficient strategies to enhance overall \cd{system performance}~\cite{carabin2017review, zakharov2020energy, tang2020reinforcemenT}. 
\ar{One proposed method involves selectively activating hardware components, such as sensors and communication peripherals, $on$ or $off$ depending} on task requirements~\cite{Rappaport2016}. 
\ar{This contrasts with traditional SLAM approaches, which require a continuous stream of sensor data at a minimum rate of 10 Hz to operate effectively~\cite{cioffi2022continuous}}. These features, including battery charging decisions, can also be integrated into the decision-making algorithms of edge/cloud task planners, especially when operating over dynamic mobile networks.

The overall efficiency of SAR operations hinges on both their real-time execution and prior planning. Effective strategic planning involves optimally allocating and distributing equipment resources across the deployment area to significantly reduce mission response times. Factors such as the battery capacity of the robots, required number of robots, and the characteristics of the area (all of which are inherently influenced by the underlying mobile network's capabilities and limitations regarding coverage, bandwidth, and latency) are key to develop a plan that anticipates the mission demands.

In our previous work~\cite{online_OROS} we proposed to integrate the orchestration logic from the mobile network infrastructure and the robot domains in an \emph{online} manner, thus enabling information exchange between the robots and a centralized control-level task planner in real-time. Despite achieving promising results in mission efficiency, the outcomes of such approaches heavily depend on the initial assumptions and conditions considered.

Given the importance of such mission planning decisions, we further proposed a novel cellular-enabled robotic SAR framework~\cite{Romero24}, that enhanced state-of-the-art orchestration strategies for mobile collaborative robots by introducing a SAR mission planning phase. Specifically, we introduced a mission planning building block that was able to take into account readily available information to first responders such as the area to be explored and the number of robots available and, considering mission goals such as exploration rate and response time, provided informed decisions on the number of robots required for a mission, \cd{ which was evaluated analytically}.

\ar{The paper at hand further improves upon the mission planning by introducing the Global Navigation Satellite System (GNSS) data pipeline, extending the quadruped robot energy profiling and incorporating realistic maps to the mission planning with Gazebo\footnote{https://gazebosim.org} simulations. 
While the original framework~\cite{Romero24} demonstrated promising results in mission efficiency, its effectiveness depended heavily on the initial assumptions and conditions. In this extension, we improve the mission planning component by integrating realistic maps to cater for more complex, real-world environments. Then, we propose the use of this environment data altogether with a motion energy model based on the robot typology. In the case of quadruped robots, we have opted for extending our previous energy profiling to account for energy consumption related to slopes. 
By simulating various mission parameters, including exploration rate and response time, we can make more informed decisions about the number of robots required, further enhancing the framework's real-world applicability.}

The remainder of this paper is structured as follows.
Section~\ref{sec:SoTA} summarizes related works in the field. 
Section~\ref{sec:SARFramework} presents our SAR framework, which consists of two phases, the \textit{Mission Planning} and the \textit{Mission Execution}.
Section~\ref{sec:Planner} delves deeper into the \textit{Mission Planning}, explaining its algorithm, and inputs, with special focus on the Global Navigation
Satellite System (GNSS) data and the robot energy profile. Section~\ref{sec:Evaluation} provides an exhaustive evaluation and, finally, Section~\ref{sec:Conclusions} concludes this paper  highlighting the future work.

\section{Related Work}
\label{sec:SoTA}

The adaptability and robustness of robot devices make them a valuable asset in hazardous environments, and it is no surprise that several works in the literature already investigated the adoption of mobile robots for SAR operations~\cite{SAR_review_2007}. In unstructured environments such as post-disaster areas, key metrics like response time and area coverage depend on multiple external factors, including the exploration strategy, the collaborative multi-robot system implementation, as well as robotic energetic and hardware and communication resources~\cite{SAR_review_2020}.
Several works in the literature tackle these issues, but mostly in an independent manner. Energy consumption savings can be achieved by enabling/disabling certain robot hardware components, particularly communication modules, sensors and processing units, increasing the exploration efficiency and robot autonomy when revisiting already known areas~\cite{rappaport_2016}.  Additionally, energy savings in mobile robots can be enhanced by performing computation off-loading~\cite{chaari2022dynamic}.

Efficient robot coordination is paramount for enhancing the overall efficiency of multi-robot SAR operations. While some centralized orchestration schemes manage robot fleet path planning using edge computing and Wi-Fi technology for communication~\cite{LNORM_2023} the advent of 5G and beyond bring new opportunities.
 Some works propose a comprehensive joint 5G and robot orchestration logic~\cite{offline_OROS}, which is then experimentally evaluated in a testbed~\cite{Groshev2025}. This logic extends beyond optimal path planning, encompassing intelligent management of robot hardware (e.g., sensors, communication peripherals) and battery charging priorities using a charge station. 
The overall proposed architecture revolves around the possibility of using a dedicated 5G network by including a gNodeB 5G-New Radio base station in the SAR equipment deployment, which guarantees fast communications between the orchestrator and the robots. The results in this paper and in previous works denote that optimal performance in SAR operations is improved upon increasing the robot fleet size, while also negatively affected by the exploration area obstacle density~\cite{online_OROS}. Building on these insights, advanced orchestration frameworks have emerged that tackle intricate planning challenges. For instance, REACT~\cite{REACT} is a smart energy-aware orchestrator specifically designed for complex indoor SAR missions. It optimizes exploration by dynamically adjusting robot paths and managing resources to ensure prolonged operational time and effective area coverage, leveraging real-time data exchange and cooperative algorithms.

Nonetheless, SAR performance can also be enhanced by focusing on the planning before mission execution. 
\ar{Area exploration performance is significantly influenced by the size of the deployed robot fleet~\cite{yan_2014}.}
Results derived from a 3D frontier-based multi-robot collaborative framework denote a positive increase in exploration efficiency and fleet size, up to a certain limit. This justifies the existence of an optimal resource allocation point that depends on the specific target deployment area. \ar{Furthermore, the effectiveness of area exploration and multi-robot performance is also influenced by the initial positioning of robotic teams, and need to be carefully considered during the initial planning phase~\cite{Yan_2015}.}

{However, elevation data is crucial for optimizing robot path planning efficiency in outdoor environments because it helps account for changes in terrain that can impact the robot's energy consumption and speed. Steeper inclines, for example, may cause robots to expend more energy and move slower, while downhill slopes could help conserve energy. By incorporating elevation data into the path planning process, robots can be guided along routes that minimize energy usage and improve overall mission efficiency.
\ar{
Such data is commonly obtained in advance to mission planning using dedicated sensors to generate pointcloud maps~\cite{Atas_2022, zhang2022path, wei_2024}, which is not a feasible solution for coverage planning of unknown areas. In such scenarios, topographic data, high-resolution Digital Elevation Maps (DEMs) or GNSS based data can be used for such purpose~\cite{Ganganath_2015, weber_2022,Yifan_2023,xu2024wheelbase}. Such studies focus on finding the lowest energy cost path in a graph-based algorithm, such as A*. Yet, these studies confine to the path planning of a single robot, with its dynamic characteristics in analytical simulations. In the multi robot space, DEM data is used for area coverage path planning~\cite{choi2021intelligent, Tang_2021}. Such algorithms are able to determine offline optimal path planning with physical restrictions while optimizing energy efficiency. However, such studies do not focus on resource optimization, neither consider the battery capacity constrains in the initial planning, and do not include a 3D physics simulator in their planning solutions.
}

{Optimal multi-robot energy-aware task planning requires accurate predictability of energy consumption while performing certain tasks. Thus, having an accurate energy profile while the robot is performing mission-related actions is crucial. In this context, a study investigates the energy consumption of a quadruped Unitree GO1 robot while staying idle and walking at 0.4 m/s~\cite{diogo_energy_profile_2025}. However, this study is very limited to general consumption, without providing specific components consumption. Other studies have examined the energy consumption of the same quadruped robot, analyzing its movement across various velocities as well as the contributions of its hardware and software components~\cite{Romero24}. Furthermore, in quadrupedal robotic systems, the energy consumption associated with terrain traversal might not be positively correlated to the slope angle~\cite{Kolvenbach_2022}. Instead, it is influenced by a combination of factors including gait selection (e.g., trot, running trot), slope angle and slope angle of attack, foot morphology and sinkage characteristics, as well as the robot’s mass and heading speed~\cite{fan2024review}. Moreover, terrain typology has also an impact on energy expenditure. For example, climbing a continuous slope imposes different dynamic and energetic demands compared to overcoming a vertical step, even though both result in the same elevation gain.

While existing literature includes studies on quadruped robot energy modeling, particularly for platforms like Unitree A1 and GO1~\cite{yang2019modeling, li2024dynamic}, a significant gap remains. Specifically, comprehensive real-world testing of these robots across diverse slope scenarios is lacking. Consequently, current research does not offer sufficiently extensive energy profiles that accurately characterize motion consumption alongside hardware and software mission-related energy demands in such uneven terrain. Crucially, none of the aforementioned works have considered the energy aspects in their initial resource planning evaluation, nor the impact energy savings might have in the resource allocation prior to mission execution. This absence of granular, real-world energy data hinders accurate mission planning and resource allocation, particularly for cellular-enabled robots whose operational longevity directly impacts the efficiency of mobile network resource utilization in challenging SAR environments.

\begin{figure*}[t!]
  \centering
\includegraphics[width=1\textwidth]{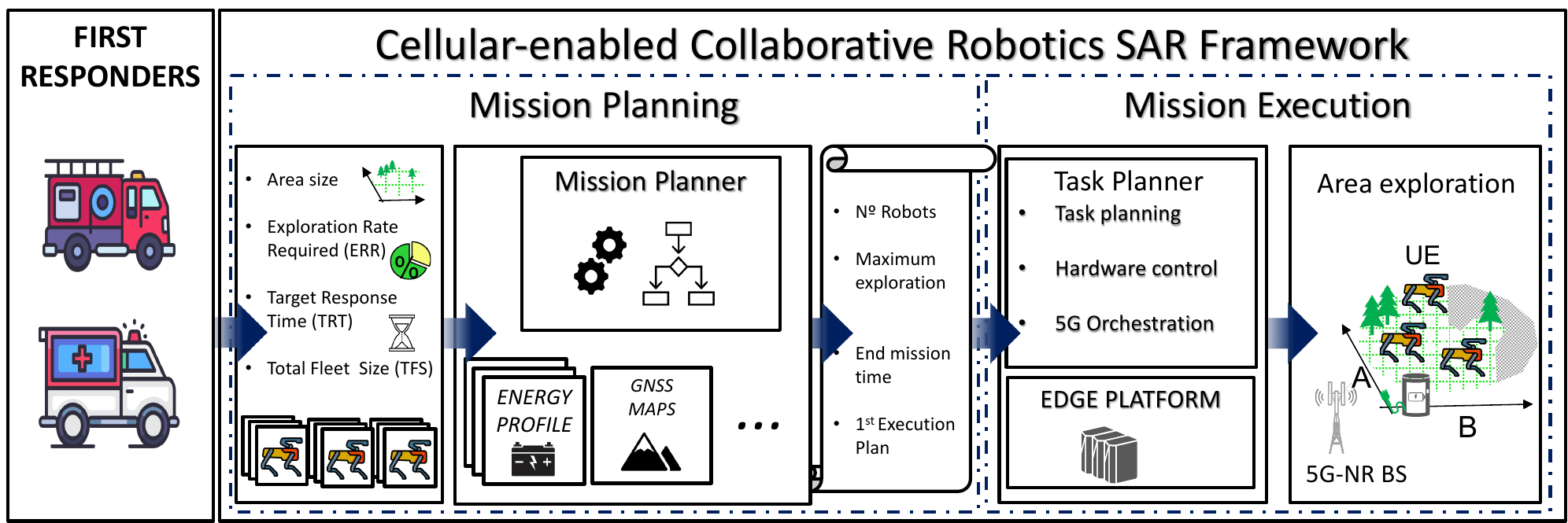}
  \caption{Cellular-enabled Collaborative Robotics Search-and-Rescue Framework.} 
  \label{fig:Framework_fig}
\end{figure*}

\section{Cellular-enabled Collaborative Robotics Search-And-Rescue Framework}
\label{sec:SARFramework}
~\cd{In this work, w}e consider scenarios where first responder teams leverage on a fleet of robots for SAR mission-critical operation\cd{s} in unknown areas.
We assume the size of the \cd{area to be explored} is known (as it may be easily estimated), and that robots \cd{can} collaborate by making use of their cellular connectivity (5G-NR). 
In order to optimize the first responders operation, we designed the SAR framework depicted in Fig.~\ref{fig:Framework_fig}.
It is composed of two main phases: the \emph{Mission Planning} phase and the \emph{Mission Execution} phase. In the following we describe them in detail.

\subsection{Mission Planning}
\label{subsec:PlanPhase}

The \emph{Mission Planning} phase is designed as an offline step preceding mission deployment. In SAR missions, response time and used equipment resources are two key factors that determine their efficiency. In general, increasing the number of robots in collaborative scenarios tends to reduce the operation time. However, the number of robots available for missions is finite and their capabilities limited by the battery capacity and consumption during operation.

\ar{A \emph{Mission Planner} is essential for determining the minimum number of robots needed to efficiently complete a mission, optimizing not only robot utilization but also indirectly supporting efficient mobile network resource allocation, while ensuring that some robots remain available for other tasks if possible. This planner operates offline, leveraging mission characteristics, the number of available robots, and prior environmental data. By analyzing these inputs, the planner predicts \emph{Mission Execution} performance and identifies the optimal fleet size needed to achieve the mission objectives.

The effectiveness of the \emph{Mission Planner} significantly improves with more accurate data and detailed prior knowledge of the mission environment. Factors such as robot specifications, energy profiles (which also includes communication energy consumption), terrain characteristics (which impact both robot mobility and signal propagation), and obstacle locations contribute to enhancing the planner’s accuracy. These considerations provide crucial insights that must be accounted for before mission deployment, and their acquisition often benefits from, or is constrained by, the capabilities of advanced mobile networks that can provide high-fidelity environmental mapping and data backhaul. \rev{Furthermore,  \emph{Mission Planning} is performed \textit{offline} prior to deployment. Therefore, it does not need to meet hard  real-time constraints.
}
}

\subsection{Mission Execution}
\label{subsec:ExecPhase}
During the \emph{Mission Execution} phase, the output of the \emph{Mission Planner} serves as a starting point, undergoing continuous updates based on real-time field data. In this phase, robots deployed in the field are expected to operate in coordination using navigation and exploration strategies to cover the target area, relaying heavily on robust, low-latency mobile network connectivity for timely command dissemination, sensor data backhaul and inter-robot communication. Additionally, using energy-aware strategies can also significantly increase the exploration efficiency.
In our work, we assume that the responsibility of ensuring an energy-aware path planning for the designated set of robots, achieved through the orchestration of their hardware and network resource utilization and task management, lies within the scope of an edge/cloud-based task planner.
The task planner is designed as a centralised high-level control entity that performs optimization decisions by performing hardware/software control, i.e., switch on/off peripherals and related drivers, as well as cellular radio resource allocation. By integrating the capabilities of activating and deactivating sensors, communication peripherals and off-loading of computation processes, the task planner provides field robots with an energy-aware coordinated task plan upon area exploration.
Multiple examples of such task planners \ar{were discussed in Section~\ref{sec:SoTA}. In this work} we assume the usage of the task planner described by~\cite{online_OROS}\cd{, \textit{OROS}, an energy-aware task planner for cellular-enabled collaborative robots. }\rev{ This task planner explicitly considers each robot's availability during its planning process and designates an exploration task as complete only upon achieving a predefined coverage threshold. The \textit{OROS} design presumes the target area is either unknown or only partially-known, thereby enabling a replanning of tasks when new obstacles are identified. However, if the obstacle is sufficiently small, the robots can traverse it using their inherent navigation strategies. In the unfortunate event a robot is lost or its battery has depleted, it will be labeled as unavailable, and its incomplete tasks will be re-included in future replanning cycles for reallocation to other available robots. Conversely, if a robot loses communication connectivity and subsequently re-establishes it, it will transition back to an available status and be integrated into the subsequent task planning process.}

\section{Mission Planner - Under the Hood}
\label{sec:Planner}

Fig.~\ref{fig:Framework_fig} illustrates the \textit{Mission Planner} within the SAR framework and presents its architecture. 
The input parameters are: i)~exploration area size, ii)~exploration rate required (ERR), i.e. percentage of the whole area to be explored, iii)~target response time (TRT), i.e. maximum time envisioned to complete a mission, and, iv) total robot fleet size (TFS). 
\ar{Then, using preloaded mission characteristics data} (e.g. robots energy profiles and \ar{GNSS data}) the \emph{Mission Planner} launches a multi-robot resource planner optimizer to find the minimum fleet size for a given mission. \cd{GNSS data provides the elevation information essential for optimizing robot path planning in these outdoor environments. By incorporating GNSS-derived elevation data, the path planning process can account for terrain variations that affect energy consumption and speed, highly affecting the operation. }

As output, the \textit{Mission Planner} provides: i)~the number of robots to be used \cd{(a subset of TFS)}, ii)~the exploration area percentage expected  \cd{(which is at least as large as ERR)}, iii)~the mission completion time \cd{(which should be TRT at most)}, and, iv)~an initial multi-robot path plan to perform during the \emph{Mission Execution}. \ar{All variables and system parameters used by the \textit{Mission Planner} are resumed in Table~\ref{tab:optParam} to allow faster referencing. } 

Algorithm~\ref{alg:MP} summarizes the \textit{Mission Planner} implementation in pseudocode. As it can be observed, 
the \textit{Mission Planner} iteratively evaluates a multi-robot resource planning problem considering an energy-aware optimization solution. At each iteration we consider the usage of an increasing number of robots (with their corresponding energy profile and battery size) and determine the amount of time required to satisfy a predetermined ERR within a given TRT. The robot fleet size is increased by one at each iteration until either the ERR and TRT requirements are met or the total available fleet size is reached with no feasible solution. 

\ar{
A detailed description of the \rsf{Multi-robot Resource Planner (\emph{RP})} optimization problem evaluated at each iteration is described in Section~\ref{subsec:optimization}. Then, Section~\ref{subsec:GNSS data} focuses on how to include GNSS data in the \textit{Mission Planner} to further improve the planning accuracy. Finally, Section~\ref{subsec:Energy Profiling} explains how robots characteristics, equipment and typology affect its energy profiling and impact on the problem accuracy.}

\begin{algorithm}[t]
\small
\SetKwInOut{Input}{Input}
\SetKwInOut{Output}{Output}
\SetKwInOut{Return}{return}
\SetKwInOut{Initialize}{Initialize}
\SetKwInOut{Procedure}{Procedure}
\Input{ $TRT, TFS, ERR$, $\mathcal{S}$ \;}
\Procedure{}
      \While{ !solved}{
        UPDATE $\mathcal{R} \subset TFS$\;
        SOLVE \textit{RP} ( $\mathcal{T}, \mathcal{R}, ERR$) \;
        GET $d_{t},e_{t,a,b}, l_{r,t,a,b}\forall t \in \mathcal{T}$\;
        \If{$\sum d_{t} \leq TRT~OR~\mathcal{R} == TFS$}{
        solved = True\;
        } 
      }
      \Output{  $\mathcal{R},  d_{t}, l_{r,t,a,b}$\;}
\caption{Mission Planner}
\label{alg:MP}
\end{algorithm}

\subsection{Multi-robot Resource Planner (RP) Problem }
\label{subsec:optimization}

Hereafter, we present our assumptions, notation and problem formulation to model the multi-robot Resource Planner problem, based on the adaptation of the problem formulation described by~\cite{offline_OROS}. 

\textbf{{Input variables.}} Let us consider a discrete set of time instants $\mathcal{T} = \{t_1, \dots, t_{|\mathcal{T}|} \}$, and a set of robots $\mathcal{R}=\{r_1, \dots, r_{|\mathcal{R}|} \}$. Each robot is equipped with a battery characterized by a maximum capacity $B_{max}$, whose charging status $b_{r,t}$ $\forall r \in \mathcal{R}$, varies over time depending on the robot activities and hardware usage. 
We assume the set of robots $\mathcal{R}$ to be deployed in an area of interest covered by mobile infrastructure for 5G connectivity. We consider the area of dimension $A \times B$ and discretize its 2D surface into a grid $\mathcal{S} =\{s_{a,b}, \forall (a,b) \in (A,B) \}$,  where each element $s_{a,b} \in \mathcal{S}$ needs to be explored.
We assume the same robot mobility approach described by~\cite{offline_OROS}, where the motion energy consumption of \ar{a} robot depends on $P_{move_{a,b,a',b'}}$. 
As mentioned before, robots exploit an existing mobile infrastructure for communications. We also consider $P_{TXa,b}$ as a variable representing the energy consumed by a robot for transmitting data, and $P_{RX}$ for receiving data.
Finally, we collect the energy consumption derived by all the cameras and sensors, as well as their processing, in the variable $P_{SEN}$. \cd{When coordinating multiple robots, it is important to consider differences in their sensors, battery life, and energy use. Our framework is built to handle this variety and can support different robot and sensor configurations. To keep things simple, we have removed the robot-specific suffix~$r$, but it can be added back if needed for specific setups.}

\begin{table*}[h!]
\caption{Model parameters}
\label{tab:optParam}
\small\centering
\begin{tabular}{l|p{8cm}}
\hline
\textbf{Parameter}	& \textbf{Definition}	\\
\hline
$\mathcal{T} = \{t_1, \dots, t_{|\mathcal{T}|} \}$		& Set of time instants; index $t$ refers to time instant $t_t$\\
$\mathcal{R}=\{r_1, \dots, r_{|\mathcal{R}|}\}$		& Set of robots $r_r$; index $r$ refers to \rsf{robot} $r_r$\\
$A \times B$                        & Geometric dimensions of the area of interest\\
$\mathcal{S} =\{s_{a,b}, \forall (a,b) \in (A,B) \}$ & Surface grid representing the area to be explored \\
$l_{r,t,a,b}$		& Binary decision variable indicating whether robot $r$ is in position $g_{a,b}$ at time instant $t$\\
$e_{t,a,b}$		& Binary variable indicating whether the unit of area $g_{a,b}$ has been explored at time $t$\\
$d_{t}$		& Binary variable indicating whether there are still areas to be explored at a time $t$\\
\rsf{$B_{max}$} & \rsf{Maximum battery capacity of robot $r$} \\
$b_{r,t}$	    & Continuous variable indicating the battery level of robot $r$ at time instant $t$, where $0 \leq b_{r,t} \leq B_{max}$ \\
$P_{move_{a,b,a',b'}}$	    & Power consumed by moving from position $g_{a,b}$ to position $g_{a',b'}$\\
$P_{RX}$	    & Power consumed for receiving \\ 
$P_{TX,a,b}$	& Power consumed for transmitting \rsf{from position $g_{a,b}$}\\ 
$P_{SEN}$	    & Power consumed by activating sensors, local data processing and on-robot computing infrastructure\\
$\kappa$	    & Exploration Rate Required (ERR) at time $t_{|\mathcal{T}|}$\\
$\mu$ & Rolling resistance coefficient \\
$m$ & Robot mass \\
$g$ & Gravity constant \\
$\theta$  & Slope angle \\ 
$v$ & Robot velocity \\
\hline
\end{tabular}
\end{table*}

\textbf{Decision variables.}
Let $d_{t}$ be a binary variable that determines whether there are still areas to be explored at a certain time $t \in \mathcal{T}$. In fact, to keep track of the multi-robot exploration, we introduce $e_{t,a,b}$ as a binary variable indicating whether the area unit $s_{a,b}$ has been already explored at time $t \in \mathcal{T}$.  Additionally, $l_{r,t,a,b}$ is a binary decision variable to control the robot mobility, its value gets positive if the robot $r$ is at position $s_{a,b}$ at time instant $t$. 

\textbf{Constraints.} Since the algorithm needs to guarantee that the ERR is satisfied, we need to enforce that the total explored area in the last time step \cd{($t_f \in T$)} is at least the corresponding ERR (i.e., the percentage $\kappa$ of the total area $|AB|$). For this purpose, we include the following constraint:

\begin{equation}
    \sum_{(a,b) \in (A,B)} e_{t_f,a,b} \geq \kappa |AB|.    
    \label{eq:exp_final}
\end{equation}

We ensure that each robot $r \in \mathcal{R}$ can only be in one place in every time instant  $t \in \mathcal{T}$:
\begin{equation}
	\label{eq:const7}
	\sum_{(a,b) \in (A,B)} l_{r,t,a,b} = 1  \quad \forall r \in \mathcal{R}, \forall t \in \mathcal{T},
\end{equation}
\noindent and with the following constraint we also ensure that the robots only move between neighbouring areas, or stay in the same position:

\begin{gather}
	l_{r,t+1,a,b} \leq l_{r,t,a,b} + l_{r,t,a-1,b} + l_{r,t,a+1,b} + l_{r,t,a,b-1} + \notag \\ 
	  l_{r,t,a,b+1} + 	l_{r,t,a-1,b-1} + l_{r,t,a+1,b+1} + l_{r,t,a-1,b+1} + \notag \\ 
 l_{r,t,a+1,b-1}  \quad \forall r \in \mathcal{R} , \forall t \in \mathcal{T}, \forall (a,b) \in (A,B).   \label{eq:const8} 
\end{gather}

In order to keep track of the exploration progress among multiple robots, if any robot $r \in \mathcal{R}$ visited 
an area unit $s_{a,b} \in (A, B)$ at some earlier time, or if it is exploring such area unit at the current time $t$, $s_{a,b}$ becomes explored at time $t$ and we update the variable $e_{t,a,b}$ accordingly.
\begin{equation}
	\label{eq:const3}
	e_{t,a,b} \leq 	e_{t-1,a,b} + \sum_{r \in \mathcal{R}}  l_{r,t,a,b}  \quad \forall t \in \mathcal{T}, \forall (a,b) \in (A,B),
\end{equation}
\begin{equation}
	\label{eq:const4}
	e_{t,a,b} \geq 		e_{t-1,a,b}  \quad \forall t \in \mathcal{T},    \forall (a,b) \in (A,B),  
\end{equation}
\begin{equation}
	\label{eq:const5}
	|\mathcal{R}| e_{t,a,b} \geq 		\sum_{r \in \mathcal{R}}  l_{r,t,a,b}   \quad \forall t \in \mathcal{T},    \forall (a,b) \in (A,B).  
\end{equation}

In order to update the decision variable $d_{t}$, according to the explored area at every time instant, we include the following constraint:
\begin{equation}
	\sum_{(a,b) \in (A,B)} e_{t,a,b} \geq \kappa (1 - d_t)  |AB|  \quad \forall t \in \mathcal{T}.
 \label{eq:act_d}
\end{equation}

Finally, as mentioned before, we assume the mobility consumption to be included in $P_{move_{a,b,a',b'}}$ and mainly dependent on the robot velocity \cd{and terrain characteristics (i.e., the terrain elevation, as will be described in Section~\ref{subsec:GNSS data})}. For the robot communications, we assume that the robot can always receive data consuming $P_{RX}$. During data transmission, the consumed power $P_{TX,a,b}$ depends on the distance to the base station. If a robot has never been in an area unit, its sensors, camera, processing units and transmission elements should be active. However, in order to reduce the energy consumption\cd{, and following the approach of~\cite{online_OROS}}, if the robot is in an already explored area, we consider the possibility to turn them off to save energy. Taking this into account, our algorithm updates the expected battery level $b_{r,t+1}$ as:

\begin{gather}
	b_{r,t+1} = b_{r,t}  - P_{RX} - \notag \\ 
	 \sum_{(a,b) \in (A,B)} \sum_{(a',b') \in (A,B)} l_{r,t,a,b} \times l_{r,t+1,a',b'} \times  P_{move_{a,b,a',b'}}  \notag \\ 
	- P_{SEN}  \times  \sum_{(a,b) \in (A,B)} (1 - e_{t,a,b}) \times l_{r,t+1,a,b} -  \label{eq:const9a} \\  
	\sum_{(a,b) \in (A,B)}   P_{TX,a,b} \times (1 - e_{t,a,b}) \times l_{r,t+1,a,b} \notag \\ \quad \forall t \in \mathcal{T} , \forall r \in \mathcal{R}. \notag
\end{gather}

%

\textbf{Objective.}
To increase the chances of detecting and assisting a target \cd{victim} in an unknown \cd{and unexplored} area \cd{the time required to explore the target area needs to be minimized:}

\begin{equation}
	\label{eq:scala_obj_g2}
	\min  \sum_{t \in \mathcal{T}}  d_{t} 
\end{equation}

To sum up, the overall problem formulation of our multi-robot resource planner can be summarized as: 

\noindent \textbf{Problem}~\texttt{RP ($ \mathcal{T}, \mathcal{R}, \kappa$)} :
\label{prob:RP}
\begin{flalign}
  \quad\quad & \text{min} \sum_{t \in \mathcal{T}}  d_{t}  \nonumber & &\\
  \quad\quad & \text{subject to:} \nonumber\\
  \quad\quad & \quad\quad (\ref{eq:exp_final}) (\ref{eq:const7}) (\ref{eq:const8}) (\ref{eq:const3}) (\ref{eq:const4}) (\ref{eq:const5})  (\ref{eq:act_d}) (\ref{eq:const9a}); \nonumber
\end{flalign}

\rev{Finally, we discuss the computational complexity and scalability of the proposed \textit{RP} formulation. The optimization problem is a mixed-integer linear program (MILP) with binary decision variables $d_t$, $e_{t,a,b}$, and $l_{r,t,a,b}$. The total number of binary variables grows proportionally to the number of robots, time instants, and grid cells, i.e., on the order of $|\mathcal{R}||\mathcal{T}||\mathcal{S}|$, where $|\mathcal{S}| = |A||B|$ denotes the number of area elements in the discretized mission region. As for any MILP with binary variables, the worst-case computational complexity of solving the \textit{RP} grows exponentially with the number of binary variables, and therefore increases rapidly with larger fleets, longer planning horizons, or finer grid discretisations. The \textit{RP} structure adopted here is consistent with our previous work on energy-aware joint orchestration of 5G and robots~\cite{Groshev2025}, where a similar MILP formulation was analysed in detail and shown to be tractable for scenarios of comparable scale.

The \textit{Mission Planner} in Algorithm~\ref{alg:MP} uses this \textit{RP} solver in an iterative fashion. At iteration $k$, it solves one \textit{RP} instance for a candidate fleet size with $|\mathcal{R}| = k$, and in the worst case it may solve up to $\mathrm{TFS}$ instances, from $k = 1$ up to $k = \mathrm{TFS}$. Hence, the worst-case complexity of the overall Mission Planner grows linearly with $\mathrm{TFS}$, on top of the exponential dependence on $|\mathcal{R}||\mathcal{T}||\mathcal{S}|$ inherited from the underlying MILP. In practice, however, the planner typically stops after only a few iterations, because feasible solutions are found for relatively small fleet sizes, and the \textit{RP} is executed offline prior to deployment. As a result, the runtime remains compatible with the pre-mission planning setting considered in this work. For significantly larger mission areas or fleets, the scalability strategies discussed in~\cite{Groshev2025} (e.g., partitioning the environment into subregions, employing moving-horizon formulations with a shorter decision window, or applying problem decomposition techniques) can be adopted to further improve scalability.

}

\ar{
\subsection{Enhancing Mission Planning through GNSS data}
\label{subsec:GNSS data}

Motion energy consumption \cd{might be} variant in \cd{real-world} terrains, specifically in \cd{uneven} rural outdoor environments, \cd{where} terrain roughness, slopes and non-structured obstacles like trees pose additional restrictions to the robot's energy efficiency. Outdoor environments might change over time but with a much lower frequency than indoor environments do, thus, relevant GNSS data can be used to map the mission planning closer to reality. From GNSS data it is possible to extract DEMs, estimate slopes, terrain typology (asphalted, grass-like, gravel, etc.) and even the positioning of obstacles that could potentially block the robot path (e.g. buildings or trees). In this section,
we explain the pipeline used to obtain this data and map it into motion energy values \rsf{to be considered} during the \textit{Mission Planning}, \cd{by characterizing the variable $P_{move_{a,b,a',b'}}$ described in Section \ref{subsec:optimization}}. 

\begin{figure*}[t!]
  \centering
\includegraphics[width=1\textwidth]{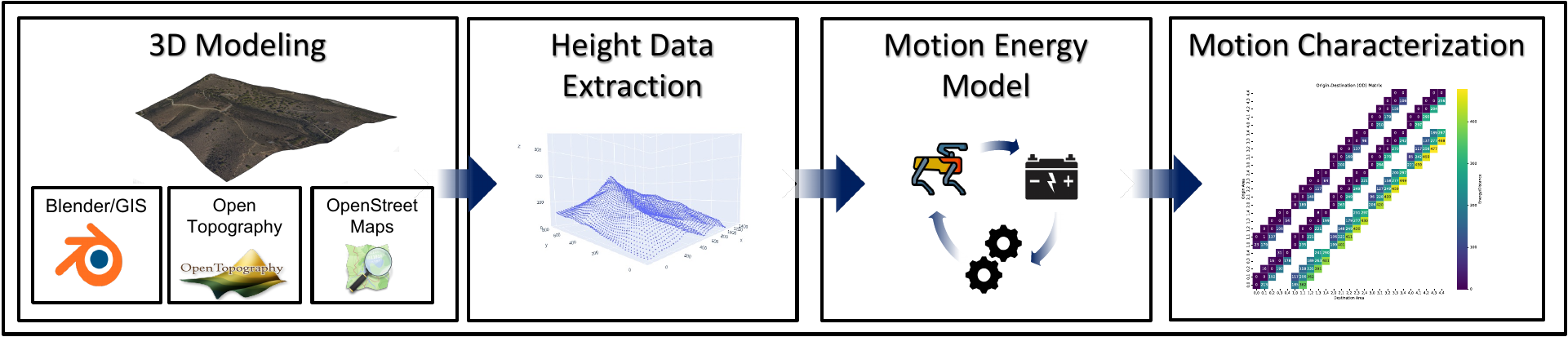}
  \caption{Enhancing mission planning through GNSS data pipeline.} 
  \label{fig:pipeline_fig}
\end{figure*}

In Fig.~\ref{fig:pipeline_fig}, we illustrate the \cd{GNSS data} pipeline, where GNSS-based 3D models are generated and their height data is extracted to compute precise energetic traversability costs using motion energy models. To generate accurate 3D representations of real outdoor terrains, we employ BlenderGIS~\cite{BlenderGIS}, an add-on for the open-source tool Blender~\cite{Blender}, in conjunction with the DEM dataset Open Topography Shuttle Radar Topography Mission (SRTM GL1) Global 30m~\cite{srtm_opentopography}. Additionally, OpenStreetMap (OSM)~\cite{OpenStreetMaps} is used to incorporate buildings into the model. We perform Height Data Extraction from the mesh geometry nodes, generating a high-resolution dataset from which slope angles are computed via normal estimation. This slope data is then converted into the estimated energy consumption for a robot with specific characteristics using a Motion Energy Model. 

Motion Energy Models depend on robot characteristics, actuators, motors efficiency, terrain friction, roughness and slope. In this work, we have considered
calculating the motion energy based on the following equation for a simplified model of wheeled robots~\cite{Minghan_2022}:
\begin{equation} E = \int (\mu (t) mg\cos (\theta (t)) + mg\sin (\theta (t)))vdt  \label{energy_equation_t}\end{equation}
\rsf{where} $\mu(t)$ is the rolling resistance coefficient in each instant $t$, $m$ is the robot mass, $g$ is the gravitational acceleration constant, $\theta(t)$ is the slope angle observed at instant $t$ and $v$ is the robot velocity. For simplicity, we assume $\mu(t)$ to be constant while predicting motion consumption along the planned path.
}

\ar{In quadrupedal robotic systems, as explained in Section~\ref{sec:SoTA}, the energy consumption depends on many variables, including robot characteristics, such as mass and heading speed. Such energy consumption might not be positively correlated to the slope angle and it might vary from one robot typology to another. In this sense, there is a concerning lack of consumption data related to commercial quadruped robots. Thus, in Section~\ref{subsec:Energy Profiling}, we present the energy profiles of a wheeled and a quadruped robot, including experimental results from tests with a real quadruped robot on different slopes.
Based on this data, we have modeled the Motion Energy Model of a quadruped robot using linear regression on power consumption over slope angle.

Finally, we do $P_{move_{a,b,a',b'}}$ Motion Characterization by calculating the motion energy consumption (E) of each robot for each viable trajectory, in our case, when moving from area $(a,b)$ to $(a',b')$ considering the slope encountered along a direct planned path.

}

\subsection{Wheeled vs Quadruped Robots - Energy Profiling}
\label{subsec:Energy Profiling}
A key aspect to be taken into account for achieving an accurate \emph{Mission Planning} is the energy profile of the robots. \cd{This includes analyzing their energy consumption patterns, battery capacity, and efficiency during different mission phases. By understanding how each robot utilizes energy for tasks such as navigation, sensing, and communication, the \textit{Mission Planner} can better optimize robot path planning and hardware utilization.} For this reason, in this subsection we focus on analyzing the energy profile of two of the most commonly robot types used in SAR operations: wheeled and quadruped robots.  
Fig.~\ref{fig:resources} depicts examples of representative wheeled~\cite{leo_rover} and quadruped robots~\cite{go1}.

\begin{figure}[t!]
     \centering
     \begin{subfigure}{0.40\columnwidth}
         \centering
         \includegraphics[width=0.9\textwidth]{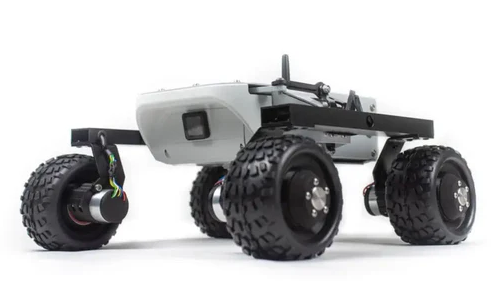}
         \caption{\small Wheeled robot~\cite{leo_rover}}
         \label{fig:wheeled_robot}
     \end{subfigure}
     \hspace{2mm}
     \begin{subfigure}{0.40\columnwidth}
         \centering
         \includegraphics[width=0.6\textwidth]{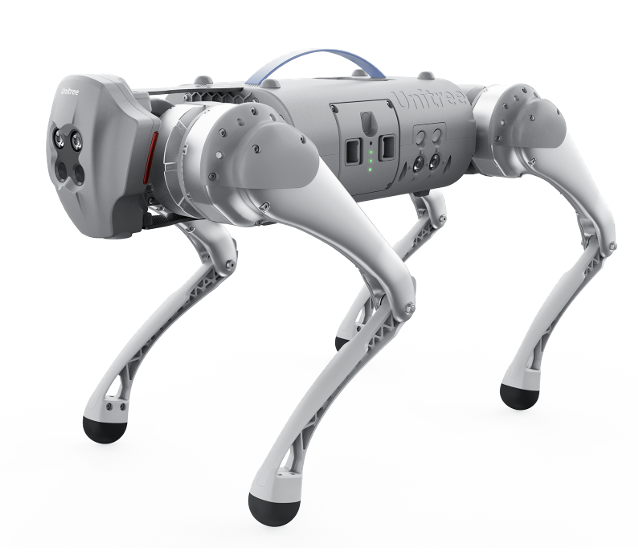}
         \caption{\small Quadruped robot~\cite{go1}}
         \label{fig:quadruped_robot}
     \end{subfigure}
     \caption{Robots 
     evaluated in the SAR Framework.}
     \label{fig:resources}
\end{figure}

Terrain-adaptability, motion speed and task-related energy consumption are key factors to be considered when choosing the mobile robots in real-world scenarios, which should be carefully evaluated upon mission planning. 
While detailed energy profiling results exist for cellular-enabled wheeled robots, e.g.~\cite{online_OROS}, our previous work~\cite{Romero24} provided an initial energy profile for quadruped robots. However, this initial work, like other studies in the literature described in Section \ref{sec:SoTA}, lacked extensive real-world testing across diverse slope scenarios. Consequently, a comprehensive energy profiling that accurately characterizes the motion consumption alongside hardware and software mission-related energy demands in such uneven terrain has remained elusive. Thus, to develop a detailed energy model for both wheeled and quadruped cellular-enabled robots within our SAR framework, we acquired a Unitree GO1 EDU robot\footnote{https://www.unitree.com/go1} and performed our own extensive \cd{energy }profiling. The results are summarized \rsf{below}.

The Unitree GO1 is a walking robot equipped with a Raspberry Pi serving as the main CPU, supplemented by an array of three additional NVIDIA Jetson Nano units. Communication capabilities are facilitated through WiFi, Bluetooth, and a cellular peripheral. 
The sensor suite of the robot comprises 5 pairs of cameras and 3 ultrasound sensors, with an additional feature being the inclusion of a 3D LiDAR that can be mounted on top the robot. Furthermore, Simultaneous Localization and Mapping (SLAM) using the LiDAR as well as human recognition through the camera feed can be performed. Notably, the Raspberry Pi perpetually powers the WiFi hotspot, while the cameras and ultrasound sensors rely on the Nano processors to which they are connected. Bluetooth, in contrast, remains in a dormant state until a new signal is received, rendering its power consumption negligible.

Table~\ref{tab:GO1_energy} presents a comprehensive breakdown of the obtained energy consumption associated with the Unitree GO1 EDU robot. Each row shows the \rsf{result} obtained when analysing the power consumption of independent robot components and motions. Tests have been performed averaging power consumption during a complete discharge of the 4500~mAh battery. For these measurements, \textit{unitree\_legged\_sdk} and \textit{unitree\_ros\_to\_real} ROS packages have been used to communicate through User Datagram Protocol (UDP) to the controller, which publishes the robot state data, including the battery state\footnote{Datasets will be available upon acceptance.}\rev{, recorded at a frequency rate of 10 Hz. Battery state data is provided from the already integrated Battery Management System of the polymer lithium-ion battery used.} The table shows the results upon evaluating the consumption when enabling/disabling non-critical robot components (i.e., cellular communications, cameras, or processors). Power consumption has been determined by comparing it to a baseline of an idle standing robot state.
\begin{table*}[t!]
\caption{Power consumption breakdown of the GO1 robot.}
\label{tab:GO1_energy}
\small\centering
\begin{tabular}{llc}
\toprule
\textbf{Category} & \textbf{Consumption Element} & \textbf{Avg. Consumption (W)} \\
\midrule
\multirow{4}{*}{Components} 
    & Cellular Peripheral & 15.77 \\
    & Cameras and Nano Proc. & 19.25 \\
    & Human Recognition & 29.38 \\
    & 3D LiDAR and SLAM & 56.84 \\
\midrule
\multirow{8}{*}{Mobility} 
    & Idle Down & 21.62 \\
    & Flexing Down & 75.79 \\
    & Idle Up & 80.33 \\
    & Flexing Up & 93.14 \\
    & Walking Circles 0.76 rad/s & 73.86 \\
    & Walking 0.5 m/s & 53.26 \\
    & Walking 1 m/s & 142.95 \\
    & Walking 2 m/s & 211.22 \\
\bottomrule
\end{tabular}
\end{table*}
As can be seen, the main \ar{consumption is due to} the use of SLAM techniques with a 3D LiDAR mounted on top of the robot. Similar consumption is observed when human recognition features (which uses the \textit{NVIDIA-AI-IOT trt\_pose}) are combined with the cameras.
The bottom section of the table denotes the results of the robot mobility tests in a flat terrain. We have considered two possible idle states: up (standing) and down (laying). The results show that in a standing position the robot consumes four times more energy than laying. Considering that it takes about 1~s for the robot to transit from up to down, and viceversa, the idle pose transition results denote that the robot can save energy by laying down in idle times longer than 2.14 s.
Four additional tests, driving straight at three different speeds and circling, have been performed. The results, which are exclusively related to robot motion, denote that energy consumption tends to increase proportionally to the robot speed.

\ar{
Additionally, we have tested the Unitree GO1 robot in different slope scenarios until battery depletion to obtain the Motion Energy Model related to terrain unevenness described in Section~\ref{subsec:GNSS data}. Fig.~\ref{fig:boxplot_GO1_slopes} shows the power consumption results of using a GO1 robot under different slopes
scenarios at a speed of 1 m/s. The results denote that for such slope scenarios a positive correlation exist between energy consumption and slope angle using a Unitree GO1 robot. Table~\ref{tab:power_slope_scenarios} shows the average power consumption results of such tests.}

\begin{figure}[t!]
     \centering
         \includegraphics[width=0.75\columnwidth]{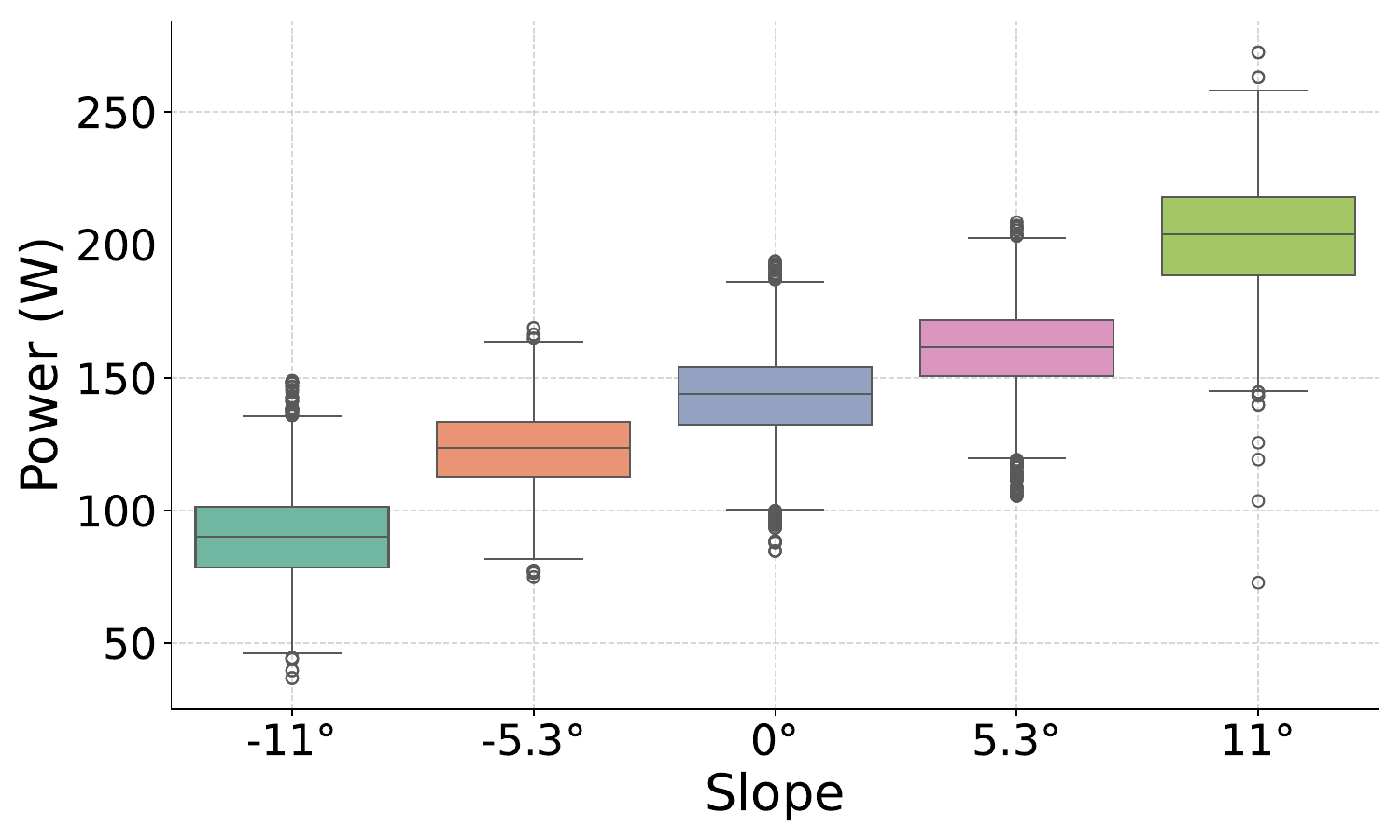}
     \caption{GO1 robot power consumption over different slopes scenarios until battery discharge}
     \label{fig:boxplot_GO1_slopes}
\end{figure}

\begin{table*}[t!]
\caption{GO1 robot average Power Consumption by slope scenario at 1 m/s}
\small\centering
\begin{tabular}{l c}
\hline
\textbf{Slope Scenario} & \textbf{Power Consumption (W)} \\
\hline
-11º    & 91.01 \\
-5.3º   & 122.87 \\
0º       & 142.95 \\
5.3º     & 160.35 \\
11º      & 203.38 \\
\hline

\end{tabular}
\label{tab:power_slope_scenarios}
\end{table*}

Table~\ref{tab:GO1_OROS_energy} summarizes the energy profiling of both types of robots, taking the values from~\cite{online_OROS} for the wheeled one and our own Unitree GO1 EDU profiling for \rsf{the quadruped one}. \rev{Measurements have been averaged from single test runs, over battery depletion.}  We compare the power consumption related to cellular communications for the reception and transmission of data, considering that both robots use similar technologies. As for sensing, consumption is related to the use of cameras, LiDAR sensor, SLAM processes and the processor. \rev{In both cases, 3D mapping is considered using Grid Map C++ library~\cite{GridMap}, with the same commercial 3D LiDAR\footnote{\url{https://gexcel.it/images/soluzioni/Velodyne/97-0038-Rev-N-97-0038-DATASHEETWEBHDL32E_Web.pdf}} guaranteeing the same sensing payloads and computational loads.}
Robot inactivity is defined as idle state, and it is the minimum consumption that robots have when performing no movement, nor using any particular hardware nor performing any action. As it can be observed, the quadruped robot has an overall higher power consumption than the wheeled robot. On the one hand, this is due to the quadruped robot having more consuming hardware components and processes than the wheeled robot. On the other hand, quadruped robots allow for more payload and are designed to explore more unstructured areas. 
Specifically, our analysis of consumption rates reveals that idle state consumption constitutes a more significant proportion of total energy usage for quadruped robots compared to wheeled robots. At the same time, wheeled robots normally have lower battery sizes than quadruped ones, emphasizing distinct energy management challenges for each typology.

\begin{table*}[h!]
\caption{Power consumption comparison.}
\label{tab:GO1_OROS_energy}
\small\centering
\resizebox{\columnwidth}{!}{\begin{tabular}{l cc cc}
\toprule
\textbf{} & \multicolumn{2}{c}{\textbf{Avg. Consumption (W)}} & \multicolumn{2}{c}{\textbf{Consumption Rate (\%)}} \\
                 & \textbf{Quadruped} & \textbf{Wheeled} & \textbf{Quadruped} & \textbf{Wheeled} \\
\midrule
Cellular Reception  & 15.77  & 4.00  & 4.75\%  & 13.97\%  \\
Cellular Transmission  & 16.72  & 4.95  & 5.04\%  & 17.28\%  \\
Camera, LiDAR, Processor  & 76.09  & 12.00  & 22.93\%  & 41.90\%  \\
Idle Up / Idle  & 80.33  & 0.29  & 24.20\%  & 1.01\%  \\
Motion (1 m/s)  & 142.95  & 7.40  & 43.08\%  & 25.84\%  \\
\midrule
\textbf{Total}  & 331.86  & 28.64  &  & \\
\bottomrule
\end{tabular}}
\end{table*}

\section{Performance Evaluation}
\label{sec:Evaluation}
~\ar{In this section, we evaluate the performance \cd{of the \emph{Mission Planner} for cellular-enabled collaborative robotics in SAR scenarios. To do so, }in Section~\ref{subsec:Scenario setup} details on the scenario setup are explained for each evaluation test performed. Section~\ref{subsec:Planning Evaluation} shows the \textit{Mission Planner} performance \cd{evaluation} results \cd{when} using the two types of robots \cd{(i.e., when considering their specific energy profiles)}. Then, in \cd{Section~\ref{subsec:Uneven evaluation}}, we evaluate the impact of the terrain unevenness by pre-loading GNSS data to our \emph{Mission Planner} and using a realistic 3D scenario in the Gazebo simulator.
}

\subsection{Evaluation Scenario Setup}
\label{subsec:Scenario setup}
\ar{
As part of the evaluation scenario setup, we consider two different performance tests: one assessing robot typology performance through analytical simulations and another evaluating terrain typology performance using physics-based Gazebo simulations. In all \rsf{the tests}, we assume a maximum battery capacity of \ar{72 kJ for wheeled robots and 350 kJ for quadruped robots}, based on their respective specifications. As for coverage, we assume that the percentage of the explored area increases when a robot reaches a target subarea and obtains a complete view of it.

In Section~\ref{subsec:Planning Evaluation}, we analyze two different exploration area sizes (50×50 m$^2$ and 500×500 m$^2$) to account for scenarios where battery consumption is either negligible or plays a significant role. As an initial approximation to evaluate the trade-offs in the \cd{\textit{Mission Planning}} optimization problem, we do not introduce obstacles in the deployment scenarios, allowing to observe the impact of different fleet sizes under ideal conditions.

Subsequently, in Section~\ref{subsec:Uneven evaluation}, \cd{we evaluate the impact of using real GNSS data. To do so,} we examine two exploration areas of 50×50 m$^2$, comparing flat surfaces and realistic uneven terrains using the Gazebo simulator with a fleet of wheeled robots.} \ar{Moreover, we evaluate the impact of terrain unevenness on the energy consumption of quadruped robot motion, using an energy model derived from the profile described in Section~\ref{subsec:Energy Profiling}. We then compare the quadruped's performance on flat and uneven terrains with that of a wheeled robot.} \rev{Two uneven terrains are considered. Their 3D models correspond to real locations near Barcelona (Spain), with the elevation modeled as previously described in Section~\ref{subsec:GNSS data}. Fig.~\ref{fig:Barcelona} shows the geographical location and the corresponding 3D terrain model of each scenario, which are denominated Uneven Terrain Scenario A and B for clarification.
}

\begin{figure}[t!]
    \centering

    \begin{subfigure}[t]{0.45\columnwidth}
        \centering
        \includegraphics[clip, trim = 0cm 0cm 0cm 0cm, width=\textwidth]{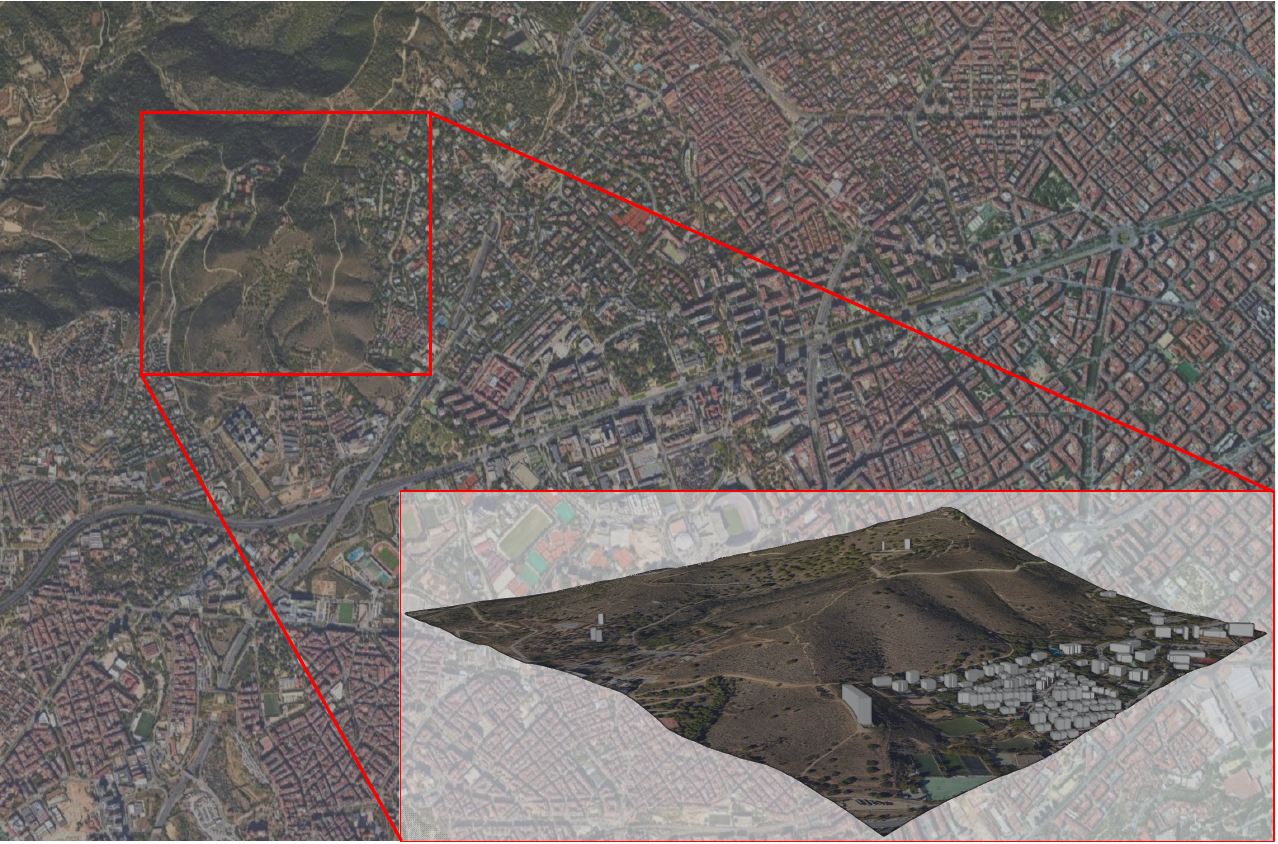}
        \caption{Uneven Terrain Scenario A location near Barcelona, Spain (41.39° N, 2.09° E).}
        \label{fig:Barcelona-a}
    \end{subfigure}
    \hfill 
    \begin{subfigure}[t]{0.45\columnwidth}
        \centering
        \includegraphics[clip, trim = 0cm 0cm 0cm 0cm, width=\textwidth]{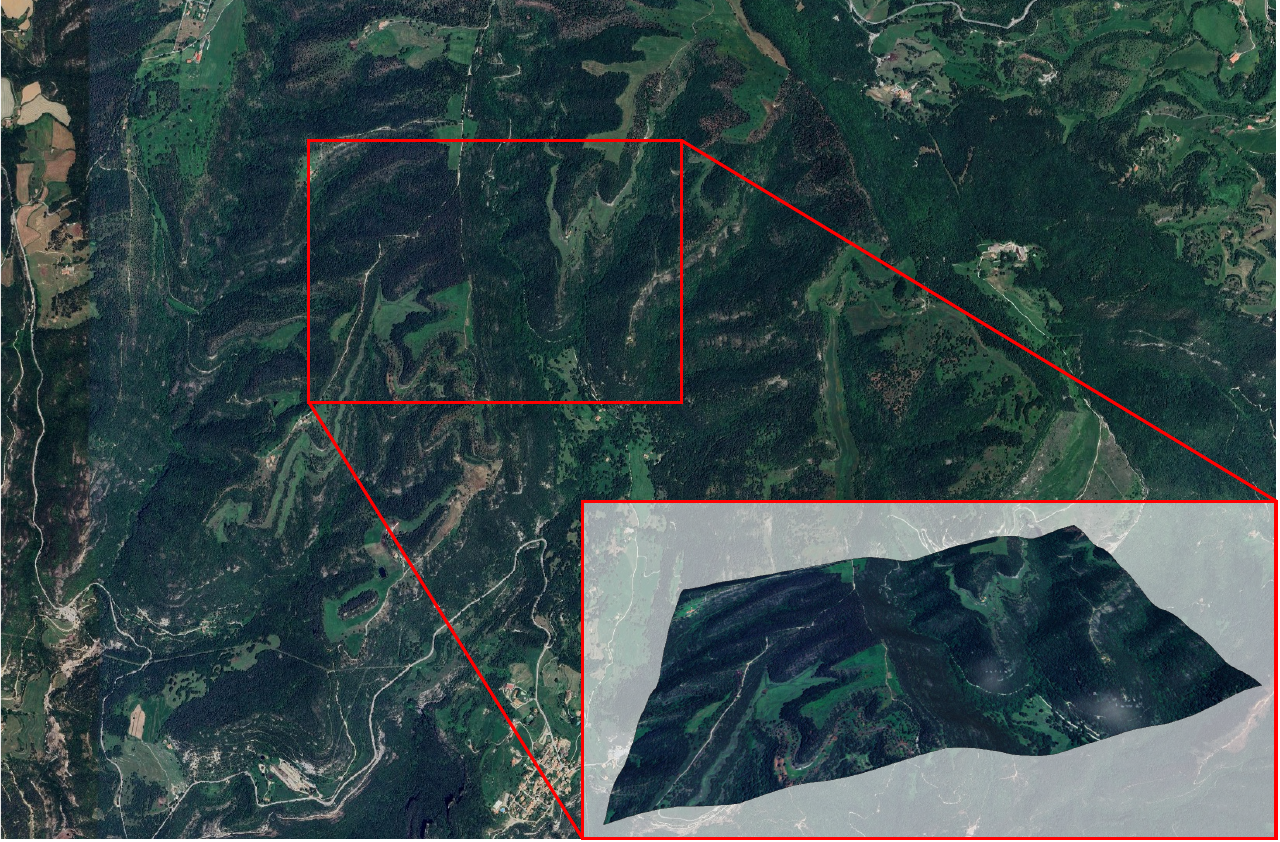} 
        \caption{Uneven Terrain Scenario B location near Barcelona, Spain (42.017° N, 2.417° E).}
        \label{fig:Barcelona-b}
    \end{subfigure}

    \caption{Visualization of the two considered uneven scenarios. Subfigures (a) and (b) illustrate the geographical location and the corresponding derived 3D terrain models for each scenario.}
    \label{fig:Barcelona}
\end{figure}

\begin{figure*}[htbp]
     \centering
     \begin{subfigure}{0.8\textwidth}
         \centering \includegraphics[width=\columnwidth]{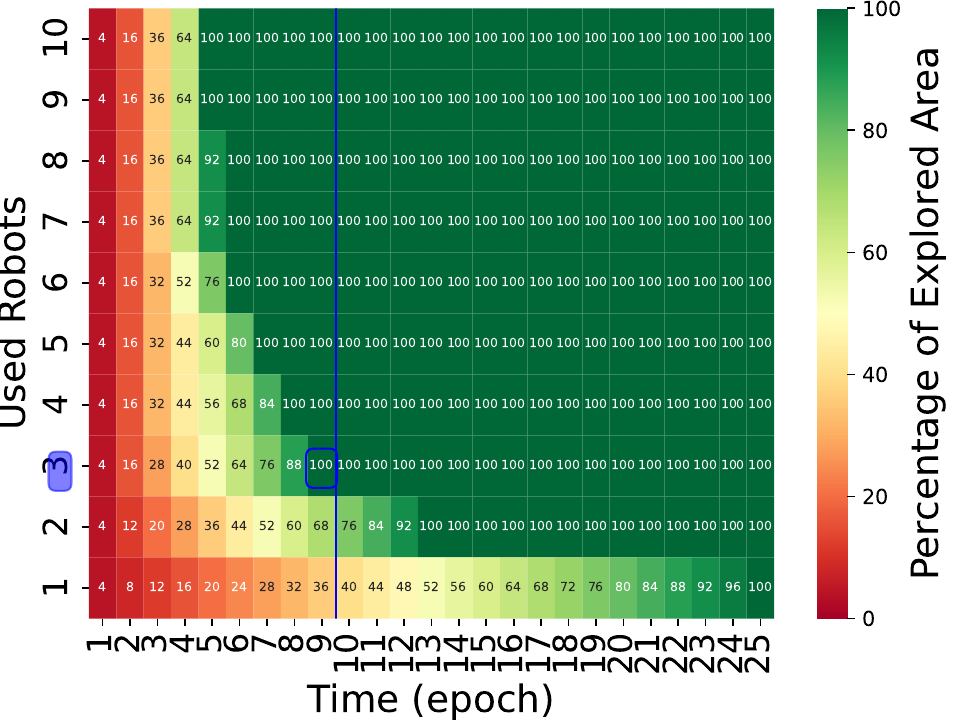}
         \caption{Wheeled robot}
    \label{fig:wheeled_50}
     \end{subfigure}
     \begin{subfigure}{0.8\textwidth}
         \centering \includegraphics[width=\columnwidth]{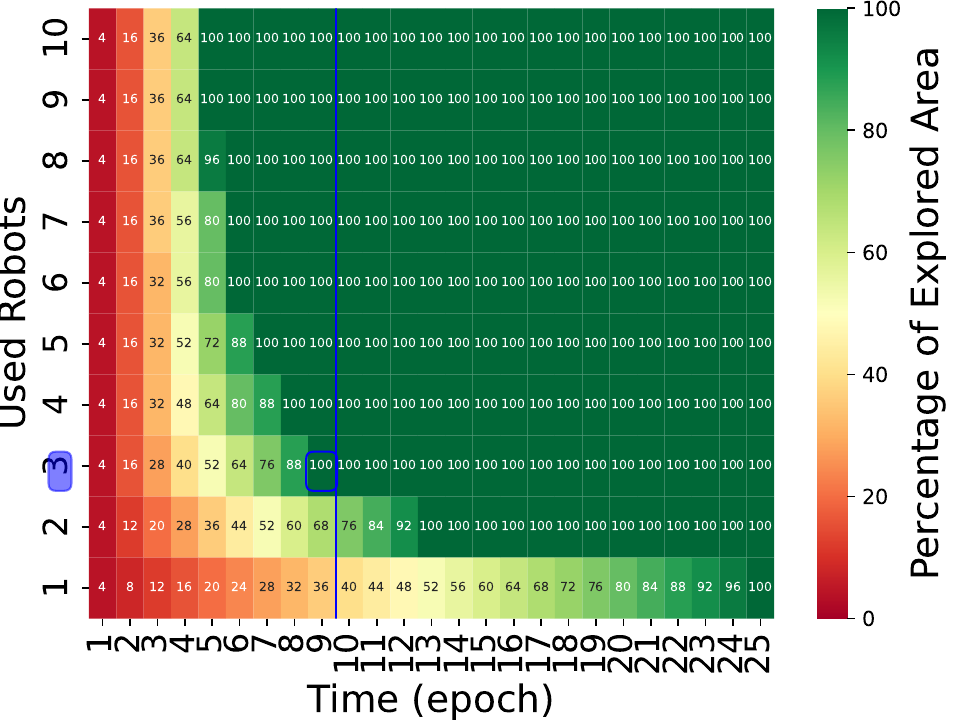}
         \caption{Quadruped robot}
         \label{fig:quadruped_50}
     \end{subfigure}
     \caption{Percentage of Explored Area per Time Epoch for a 50x50 $\mbox{m}^2$ scenario. Wheeled versus Quadruped Robots.}
     \label{fig:comparative_legg_wheel}
     
\end{figure*}
\begin{figure*}[ht]
     \centering
     \begin{subfigure}{0.8\textwidth}
         \centering
         \includegraphics[width=\columnwidth]{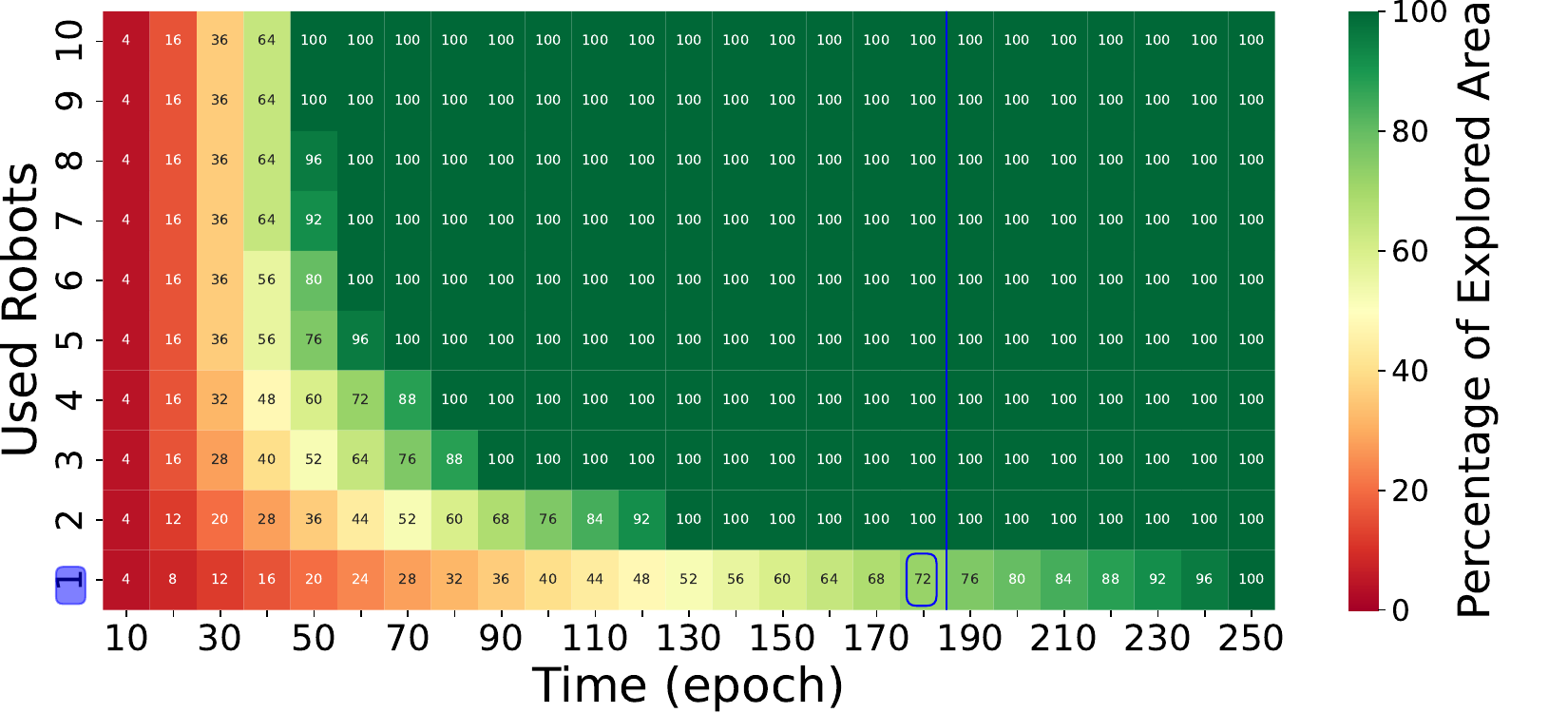}
         \caption{Wheeled robot}
         \label{fig:wheeled_500}
     \end{subfigure}
     \begin{subfigure}{0.8\textwidth}
         \centering
         \includegraphics[width=\columnwidth]{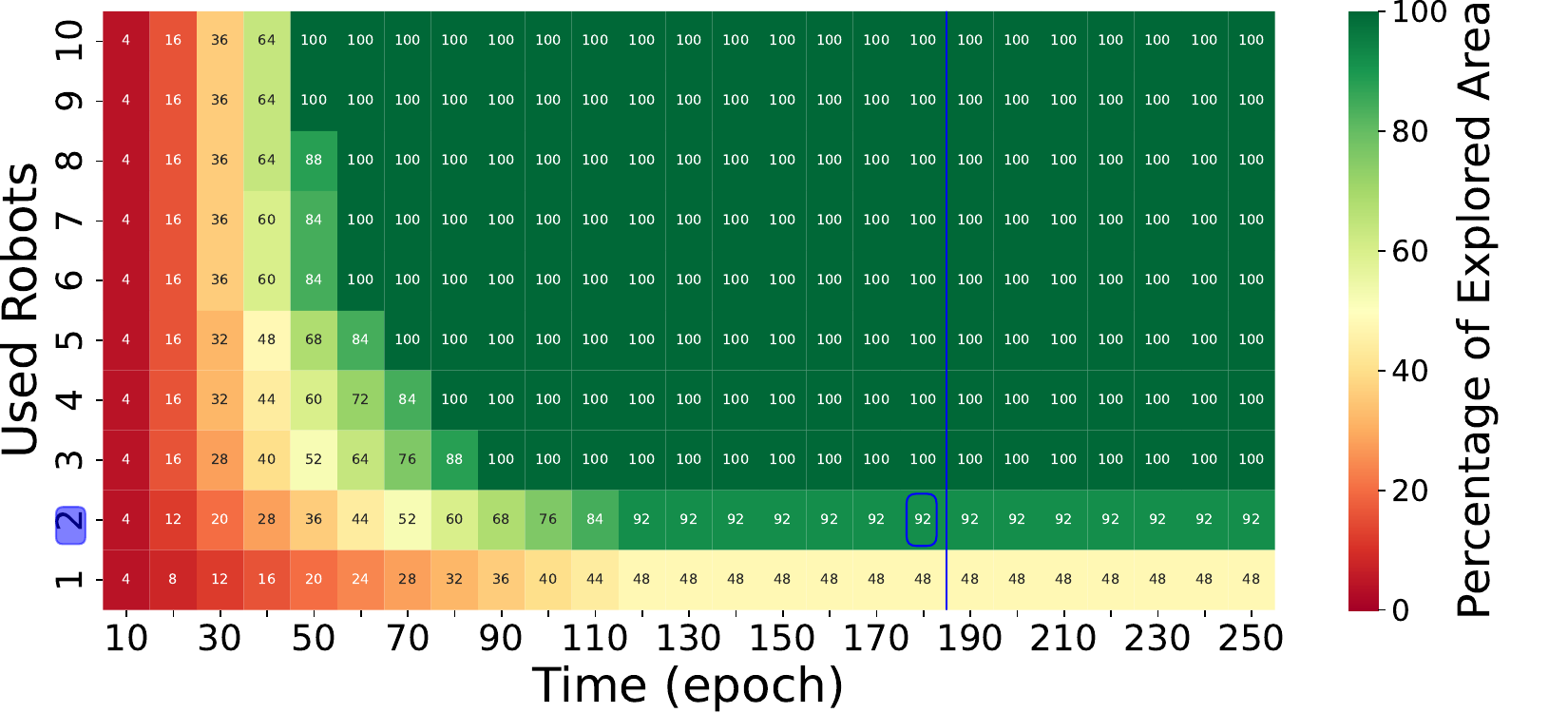}
         \caption{Quadruped robot}
         \label{fig:quadruped_500}
     \end{subfigure}
     \caption{Percentage of Explored Area per Time Epoch for a 500x500 $\mbox{m}^2$ scenario. Wheeled versus Quadruped Robots.}
     \label{fig:comparative_legg_wheel_500}
\end{figure*}

\subsection{Mission Planning Evaluation}
\label{subsec:Planning Evaluation}

In Fig.~\ref{fig:comparative_legg_wheel} we compare the performance of wheeled versus quadruped robot fleets in the~50x50~m$^2$ area moving both robots at 1~m/s. We consider the total fleet size to be 10 robots, an exploration rate required of 75\% of the total area and a target response time up to 90~s. Note that each epoch \cd{in the figure} is equivalent to 10~s in our experiments. 

The results obtained with both type of robots are similar, due to the fact that the battery capacity is sufficient to cover the totality of the area at the given 1 m/s speed. In this scenario the \textit{Mission Planner} outputs a minimal fleet size of three robots to satisfy \cd{the} target conditions, for both types of robots. 

In Fig.~\ref{fig:comparative_legg_wheel_500} the \textit{Mission Planner} has been used to evaluate the impact of a \rsf{hundred} times larger area (500x500 m$^2$).
As in the previous case, we consider \cd{a} total fleet size \cd{of} 10 robots, an exploration rate required of 70\% of the total area, a target response time up to 180 epochs given the larger size of the scenario\cd{,} and \cd{a} moving speed of 1~m/s. 

The results in this case differ between the wheeled and quadruped robots as expected, since the differences in the energy consumption between robot types become visible. Despite the fact that the larger energy consumption required by quadruped robots is compensated with a larger battery size, the impact it has in relation to its battery capacity is much greater than in the wheeled robot case. Therefore, the results of the \textit{Mission Planner} indicate that while one wheeled robot would be sufficient to meet the mission requirements, two quadruped robots would be needed in the same conditions.

\begin{figure*}[ht]
     \centering
     \begin{subfigure}{0.8\textwidth}
         \centering
         \includegraphics[width=\columnwidth]{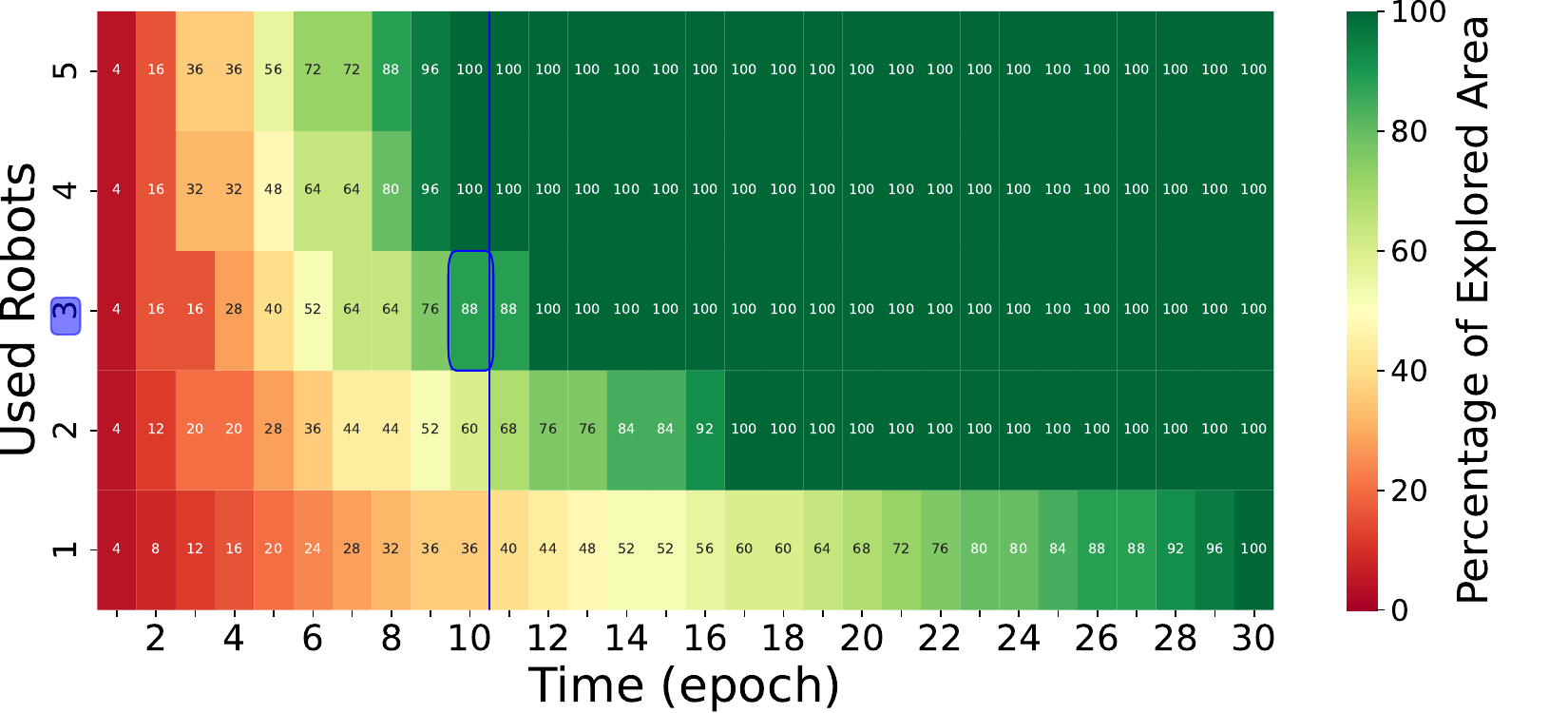}
         \caption{Flat Terrain}
         \label{fig:Flat_Terrain}
     \end{subfigure}
     
     \begin{subfigure}{0.8\textwidth}
         \centering
         \includegraphics[width=\columnwidth]{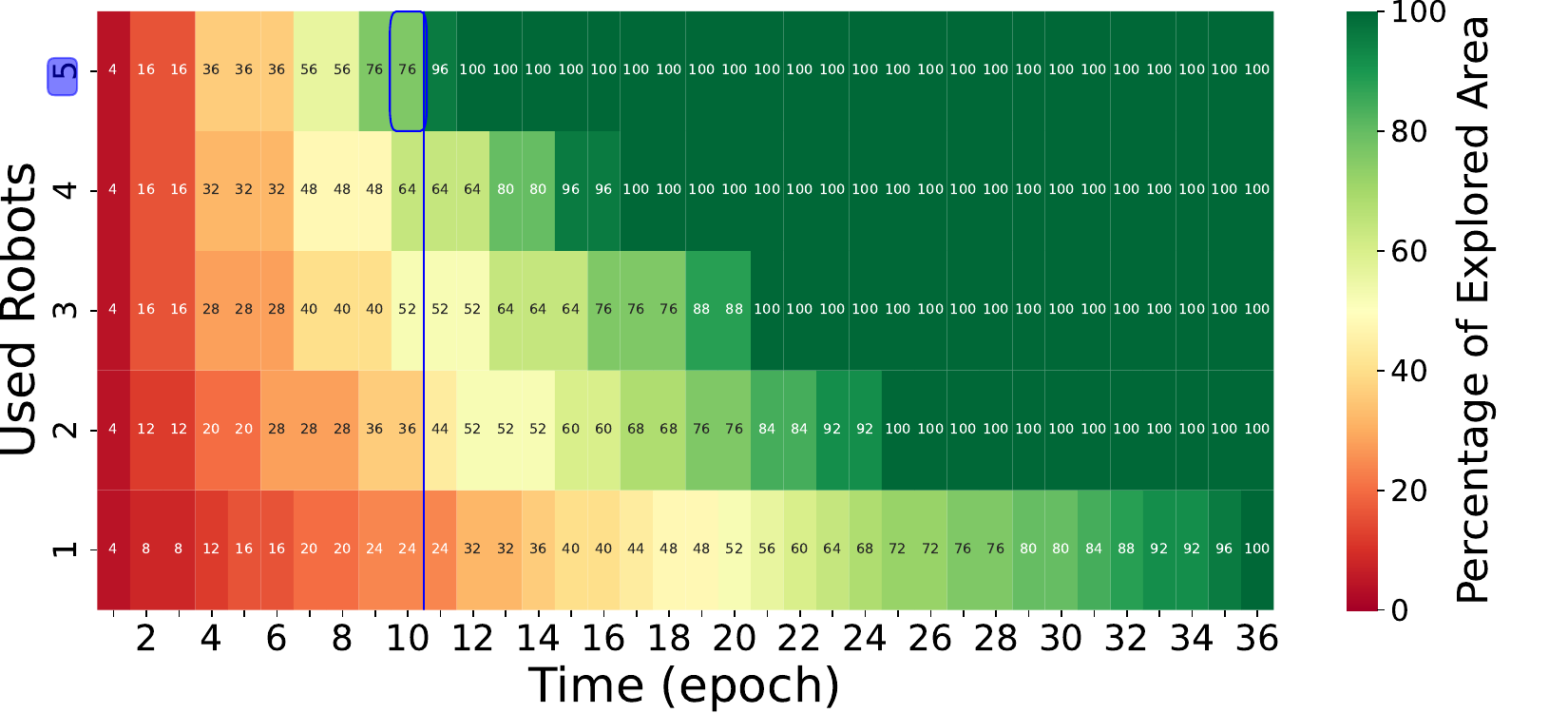}
         \caption{Uneven Terrain}
         \label{fig:Uneven_Terrain}
     \end{subfigure}
     \caption{Percentage of Explored Area per Time Epoch for a 50x50 m$^2$ scenario using a wheeled robot fleet.}
     \label{fig:comparative_flat_uneven}
\end{figure*}

\rev{Finally, we report the runtime of the $RP$ solver for the small-area case. In our implementation, solving a single $RP$ instance for the 50×50 m² scenario with 1 robot required on average 8.81 s over 100 test runs on a standard workstation: a virtual machine running Windows 10 Home with 1 Intel Processor 64 2.9 GHz and 10 GB of RAM. These runtimes show that, for mission areas of this scale, the offline Mission Planner can compute a complete exploration schedule within "a few seconds", which is fully compatible with typical pre-mission planning workflows. For larger mission areas or fleets, the same formulation can be combined with coarser grid discretisation or spatial partitioning to keep the runtime within similar offline timescales.}

\ar{
\subsection{Uneven Terrain Evaluation}
\label{subsec:Uneven evaluation}
In Fig.~\ref{fig:comparative_flat_uneven} we compare the performance of a fleet of \cd{five} wheeled robots operating in two simulated areas with different elevation characteristics, each sizing 50x50~m$^2$.  We consider an exploration rate required of 75\% of the total area, and a target
response time up to 10 epochs. The simulations were conducted in Gazebo, incorporating realistic physical and energy models with battery dynamics\footnote{Battery plugin used: https://github.com/nilseuropa/gazebo\_ros\_battery}. \cd{The robot e}nergy consumption is obtained from Table~\ref{tab:GO1_OROS_energy}, although robot motion energy consumption is computed using equation~(\ref{energy_equation_t}). We estimate a rolling resistance coefficient of \mbox{$\mu(t) = 0.1$} assuming that the robot consumes 7.4~J after traveling 1~m on a flat terrain as defined in our energy profiling. For modeling simplicity, we assume  $\mu(t)$ remains constant throughout the simulations. On the robot characteristics, each wheeled robot has a maximum linear speed of 0.4 m/s, a base weight of 6.51 kg, and an additional 1.0 kg payload for sensors. Note that each epoch
in this figure is equivalent to 35 s in our experiments. 

The results indicate that the \cd{percentage of }explored area over time follows a similar trend for both terrain types. This is because the robots' battery capacity is sufficient to offset the increased energy consumption on \cd{the} uneven terrain. In fact, the primary difference between both terrain typologies is that unstructured terrain leads to longer travel times between subareas due to elevation changes. Therefore, this has a higher impact in \cd{the r}esponse time and energy consumption. As illustrated in Fig.~\ref{fig:comparative_flat_uneven}, the output of the \emph{Mission Planner} shows that while three robots are sufficient to meet the mission requirements in a flat scenario, five robots are needed to achieve the same target response time (10 epochs) and exploration rate required (75\%) when accounting for the terrain’s unevenness. 

\begin{figure}[t!]
     \centering
         \includegraphics[width=0.8\columnwidth]{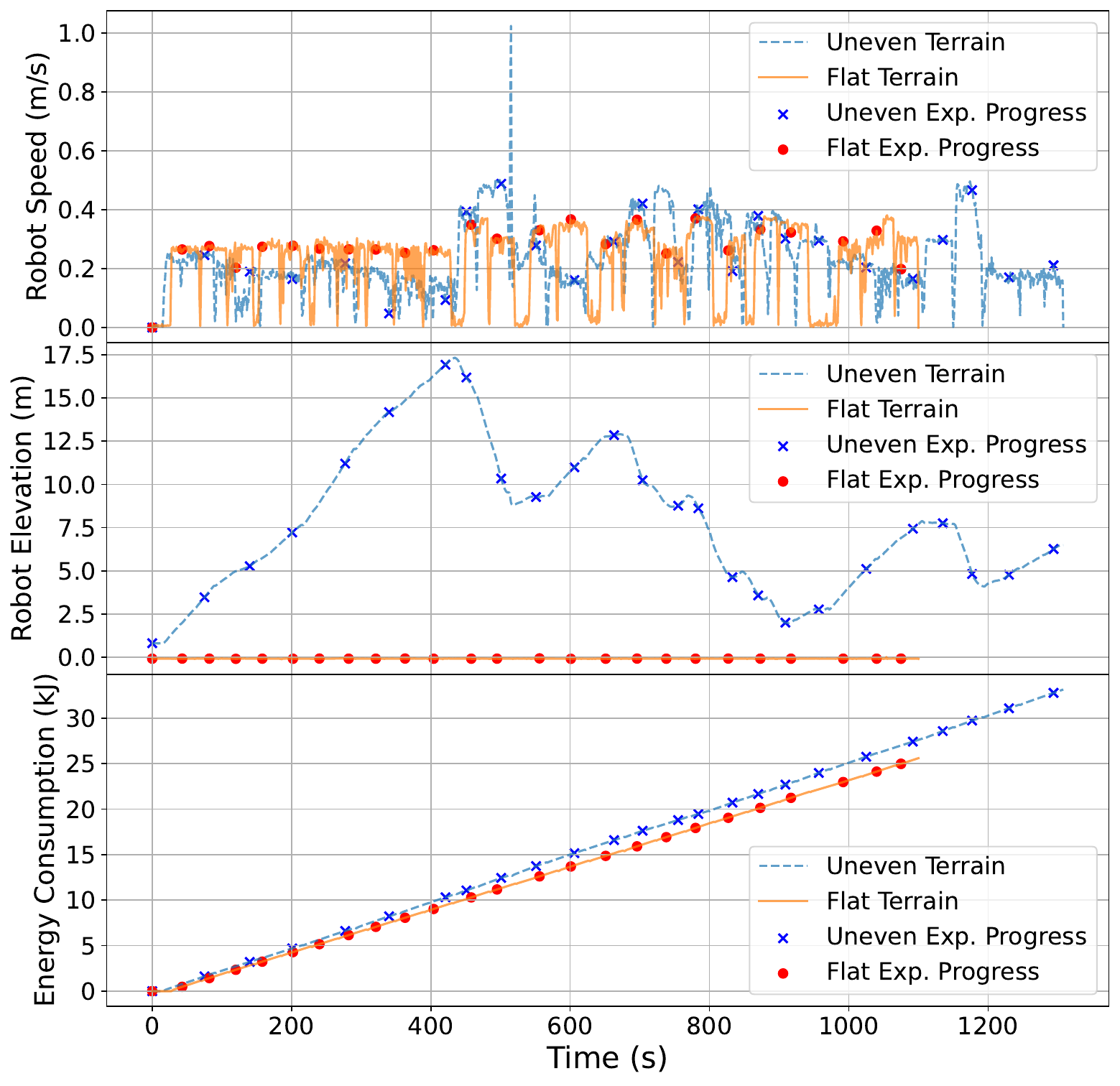}
     \caption{Terrain results using one wheeled robot with Gazebo simulations.}
     \label{fig:comparative_one_robot}
\end{figure}

The effects of unstructured terrain on exploration time progression are shown in Fig.~\ref{fig:comparative_one_robot} for the wheeled robot. Robot speed, elevation, and energy consumption are plotted over time for both terrain scenarios using a single wheeled robot. On flat terrains, \cd{speed} and energy consumption align with \cd{the }expected values. However, on uneven terrain, elevation changes impose additional energy costs, limiting speed due to motor constraints. This effect is partially counterbalanced when descending, where gravitational force aids acceleration, even surpassing the motor 0.4 m/s threshold, as can be seen at time 500~s. Additionally, exploration progression in Fig.~\ref{fig:comparative_one_robot}, which accounts for robots reaching its target subarea, is displayed as red and blue marks over the plotted data for flat and uneven terrains respectively. In a flat terrain, exploration progression has a homogeneous distribution over time while in uneven terrains its distribution is affected by the slope and its limitations on robot speed. Overall, a single wheeled robot consumes approximately 7500~J more on \rsf{the} uneven terrain compared to a flat terrain, which is about a 10\% of its total battery capacity.

\begin{figure}[t!]
     \centering
         \includegraphics[width=0.85\columnwidth]{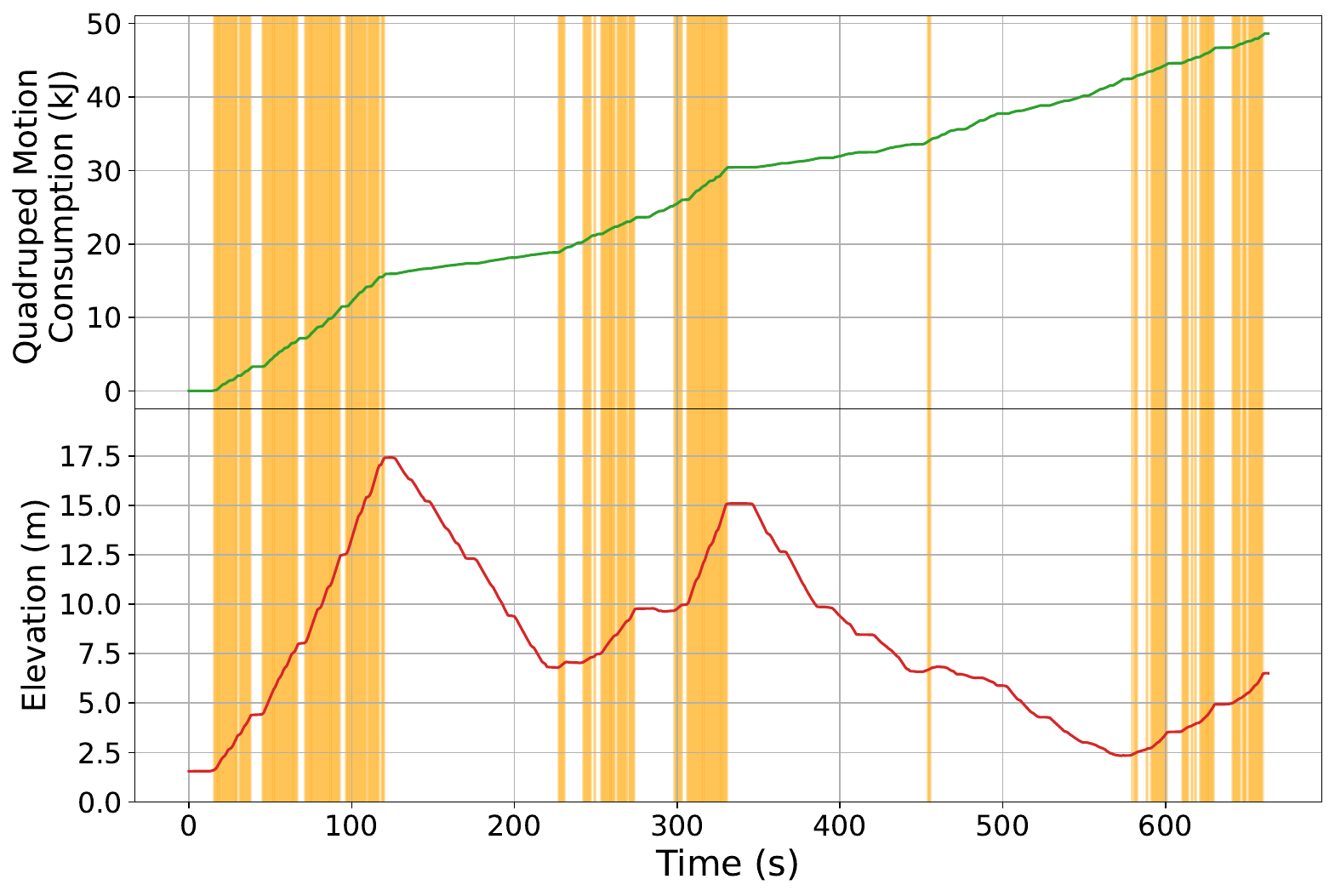}
     \caption{Aggregated motion energy consumption and elevation results using a quadruped GO1 robot with Gazebo simulations.}
     \label{fig:go1_result}
\end{figure}

In Fig.~\ref{fig:go1_result} we show the effect of considering the energy consumption related to the unevenness of the terrain using a quadruped GO1 robot in the Gazebo simulator using the \textit{Unitree\_guide} ROS package. This package is used for its robot controller, \textit{junior\_ctrl}, to simulate the Finite State Machine (FSM), the leg motion and robot state estimator. For this robot we have considered a maximum linear speed of 0.7 m/s, a base weight of 12 kg and an additional 1.0 kg payload for sensors, and its motion energy consumption is considered using data from Table~\ref{tab:power_slope_scenarios} and Table~\ref{tab:GO1_OROS_energy}. In this figure, the cumulative robot motion energy consumption and the elevation are plotted over time for a single quadruped robot. As shown in the highlighted areas, the energy consumption increases sharply during upward motion, while  being more gradual during downward and flat motion.
}

\rev{In Fig.~\ref{fig:leo_go1_motion_energy} motion energy consumption over exploration rate is plotted for a single quadruped (Fig.~\ref{fig:go1_motion_energy}) and wheeled (Fig.~\ref{fig:leo_motion_energy})  robot in three different terrain scenarios: flat, uneven terrain A and uneven terrain B. For each robot and terrain case 5 tests runs have been performed. The box plots depict the distribution of motion energy consumption in discretized exploration rates, showing the direct impact of terrain on each robot typology, which is quantified through their respective Motion Energy Model. In both robot typologies a flat terrain implies a linear increase in consumption while exploring. However, when accounting for terrain unevenness, energy consumption spikes during upward motion and decreases during downward motion. In fact, in both cases the consumption in uneven terrain B is higher, which is logical since uneven terrain A and B have a total elevation difference of 17.5~m and 30~m respectively. Upon quantifying the impact of terrain unevenness, in worst case scenarios for wheeled robots results in an additional consumption of 1.4~kJ on average, approximately a 48\% increase over the flat terrain baseline. The effect is even greater in quadruped robots, where energy consumption increases by 26.13~kJ, effectively doubling relative to the flat baseline.
}

\begin{figure*}[t!]
     \centering
     \begin{subfigure}{0.8\textwidth}
         \centering
         \includegraphics[width=\columnwidth]{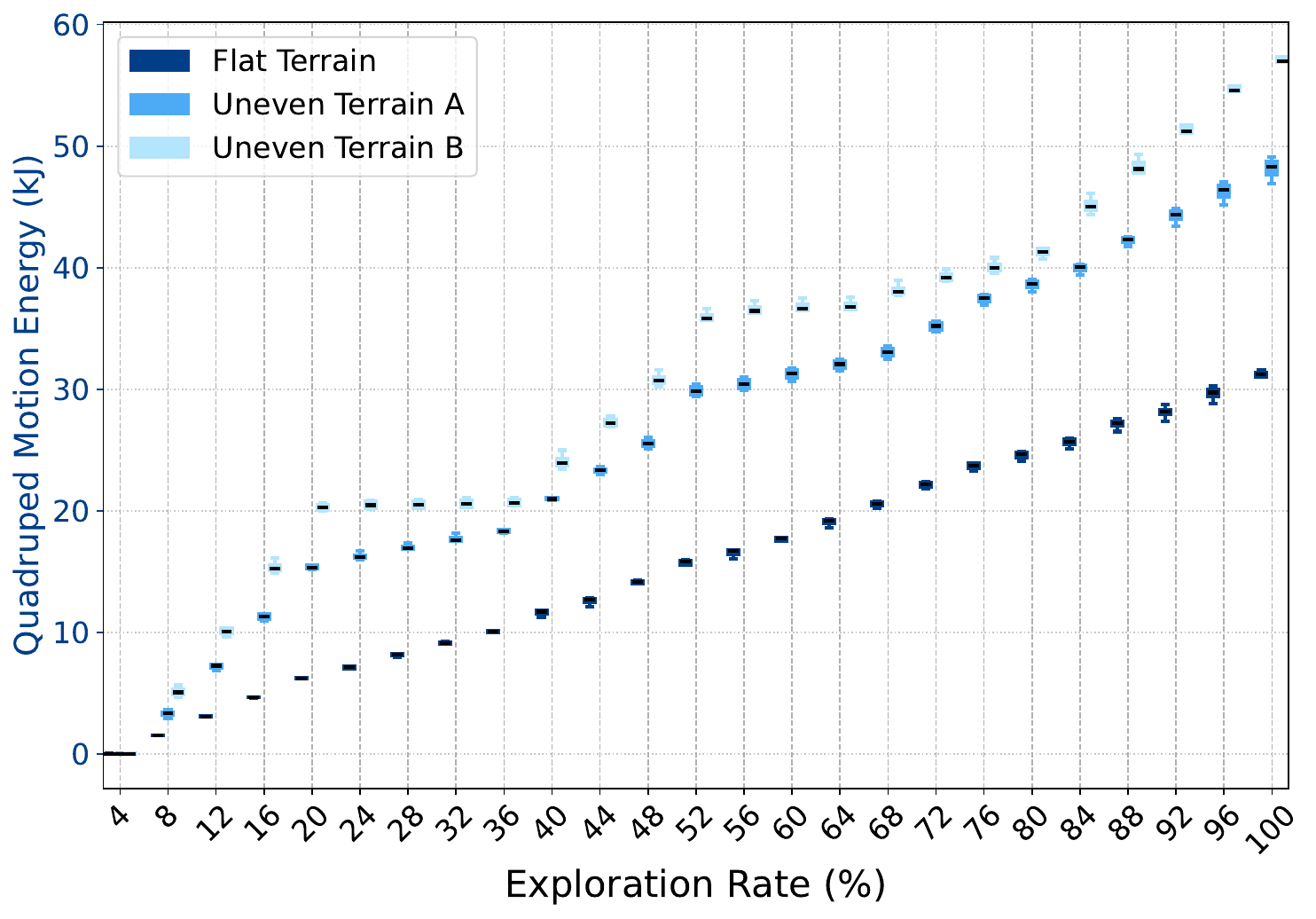}
         \caption{Single Quadruped robot motion energy consumption over percentage of explored area in flat and uneven terrains.}
         \label{fig:go1_motion_energy}
     \end{subfigure}
     
     \begin{subfigure}{0.8\textwidth}
         \centering
         \includegraphics[width=\columnwidth]{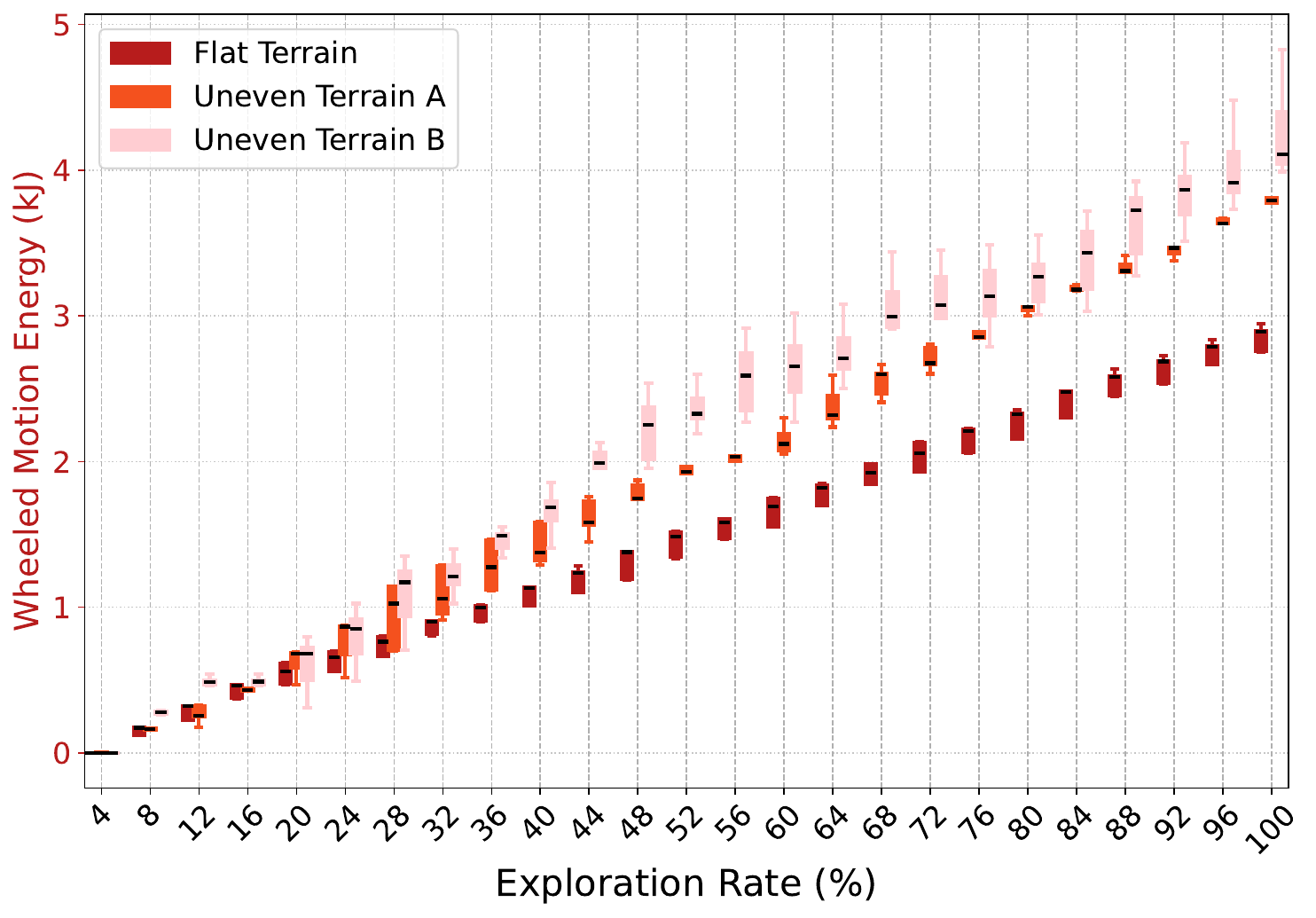}
         \caption{Single Wheeled robot motion energy consumption over percentage of explored area in flat and uneven terrains.}
         \label{fig:leo_motion_energy}
     \end{subfigure}
     \caption{Motion energy consumption over percentage of explored area of quadruped and wheeled robots in three different scenarios.  }
     \label{fig:leo_go1_motion_energy}
\end{figure*}

\newpage

\section{Conclusions and Future Works}
\label{sec:Conclusions}
In mission-critical operations, the role of cellular-enabled collaborative robot fleets in augmenting the search-and-rescue capabilities of first responders is crucial.
In this paper, we proposed a novel SAR framework for cellular-enabled collaborative robotics mission planning that, taking as input information readily available (exploration area, fleet size, energy profile, \cd{GNSS data, }exploration rate and target response time), allows first responders to take informed decisions about the number of robots needed to successfully complete a mission. Moreover, our results illustrate the trade-off when considering different types of robots (wheeled vs quadruped) with respect to the number of necessary robots, explored area and response time, underscoring how foundational planning impacts overall system efficiency, including communication load. 
\ar{We have also shown the importance of Motion Energy Models in \emph{Mission Planning}, understanding real robot energy behavior while performing in different slope scenarios. Our results demonstrated the impact of terrain information on mission performance highlighting how different surface conditions influence exploration efficiency, time response and energy consumption. While terrain information in outdoors scenarios can be acquired through GNSS data, accurate robot behavior and prediction of its energy consumption requires a deeper study.  }

Future work will consider expanding our SAR framework to further consider larger scale scenarios (\ar{e.g. in terms of bigger areas, number and heterogeneity of robots, terrain diversity, obstacles)} and input parameters available (\ar{e.g. a more detailed surface information which would determine the type of robot that can be used, higher granularity robot energy profiling, higher control granularity)}. \rev{Furthermore, we will specifically evaluate the impact of limited communications scenarios that arise from factors like adverse weather conditions or poor network signal, similar to the challenges encountered in initiatives such as the DARPA Subterranean Challenge~\cite{2019darpa}. Alongside we will study the effect of dynamic obstacles during Mission Execution, focusing on the resulting unanticipated delays and the necessity for replanning. While our current framework proposal can mitigate these issues, either by relying on availability reporting or through the robots' navigation strategies, it does not currently provide optimally efficient solutions for such scenarios. Addressing this expanded complexity will likely need the application of advanced analytical and/or machine learning solutions to ensure feasibility and optimal performance.}

Ultimately, this will enable even better informed decisions for SAR operations, further enhancing the capabilities of autonomous systems leveraging the advanced features of 5G and future 6G mobile networks, such as ultra-reliable low-latency communication (URLLC), massive machine-type communication (mMTC), and integrated sensing and communication (ISAC).

\newpage

\section*{Author contributions: CRediT}

\noindent\textbf{Arnau Romero}: Investigation, Methodology, Software, Validation, Visualization, Writing – original draft. 
\\
\textbf{Carmen Delgado}: Methodology, Conceptualization, Formal analysis, Supervision, Validation, Writing – original draft, review and editing. 
\\
\textbf{Jana Baguer}: Data curation, Investigation, Methodology, Software, Visualization. 
\\
\textbf{Ra\'ul Su\'arez}: Conceptualization, Funding acquisition, Project administration, Supervision, Writing – review and editing. 
\\
\textbf{Xavier Costa-P\'erez}: Conceptualization, Funding acquisition, Project administration, Resources, Supervision, Writing – review and editing. 

\section*{Data Availability}
Datasets will be available upon acceptance.

\bibliographystyle{elsarticle-num}
\bibliography{biblio2}

\end{document}